%% file: main.tex
\pdfoutput=1
\documentclass[11pt]{article}

\usepackage{ACL2023}

\usepackage{times}
\usepackage{latexsym}
\usepackage{graphicx}
\usepackage[T1]{fontenc}
\usepackage[utf8]{inputenc}
\usepackage{makecell}
\usepackage{microtype}

\usepackage{stfloats}
\usepackage{inconsolata}
\usepackage{amssymb}
\usepackage{subcaption}
\usepackage{amsmath}
\usepackage{cleveref}
\usepackage{bm}
\usepackage{float}
\usepackage[linesnumbered,ruled,vlined,algo2e]{algorithm2e}
\usepackage{pifont}
\usepackage{booktabs}
\usepackage{multirow}
\usepackage{enumitem}
\usepackage{algorithmic}
\usepackage{algorithm}
\usepackage{CJKutf8}

\title{Head-wise Shareable Attention for Large Language Models}

\author{Zouying Cao$^{1,2,3}$, Yifei Yang$^{1,2,3}$, Hai Zhao$^{1,2,3,}$\thanks{\, Corresponding author. This research was supported by the Joint Research Project of Yangtze River Delta Science and Technology Innovation Community (No. 2022CSJGG1400),
the Joint Funds of the National Natural Science Foundation of China (Grant No. U21B2020).} \\
$^1$Department of Computer Science and Engineering, Shanghai Jiao Tong University\\
$^2$Key Laboratory of Shanghai Education Commission for Intelligent Interaction \\
and Cognitive Engineering, Shanghai Jiao Tong University \\
$^3$Shanghai Key Laboratory of Trusted Data Circulation and Governance in Web3 \\
\texttt{\{zouyingcao,yifeiyang\}@sjtu.edu.cn, zhaohai@cs.sjtu.edu.cn} \\}

\begin{document}
\maketitle
\begin{abstract}
% Despite the remarkable capabilities of large pre-trained language models (PLMs), they suffer from huge number of parameters, which restricts their deployment on devices with limited memory. 
Large Language Models (LLMs) suffer from huge number of parameters, which restricts their deployment on edge devices. 
Weight sharing is one promising solution that encourages weight reuse, effectively reducing memory usage with less performance drop. % reducing the number of parameters with minimum performance drop to some extent
However, current weight sharing techniques primarily focus on small-scale models like BERT and employ coarse-grained sharing rules, e.g., layer-wise. 
This becomes limiting given the prevalence of LLMs and sharing an entire layer or block obviously diminishes the flexibility of weight sharing. 
In this paper, we present a perspective on head-wise shareable attention for large language models. 
We further propose two memory-efficient methods that share parameters across attention heads, with a specific focus on LLMs. % to reduce the memory usage for large PLMs. 
Both of them use the same dynamic strategy to select the shared weight matrices. 
The first method directly reuses the pre-trained weights without retraining, denoted as \textbf{DirectShare}. 
The second method first post-trains with constraint on weight matrix similarity and then shares, denoted as \textbf{PostShare}. 
Experimental results reveal our head-wise shared models still maintain satisfactory capabilities, demonstrating the feasibility of fine-grained weight sharing applied to LLMs\footnote{\url{https://github.com/zouyingcao/DirectShare}}. 
\end{abstract}

\input{1-Introduction}
\input{2-Related_Works}
\input{3-Motivation}
\input{4-Method}
\input{5-Experiments}
\input{6-Conclusion}

\section*{Limitations}
This paper primarily focuses on the head-wise weight sharing in Multi-Head Attention (MHA) block, inspired by the attention map similarity across heads. 
Although we have explored the feasibility of our proposal weight sharing strategy in the Feed-Forward Network (FFN) block, we only complete downstream evaluation on Baichuan2-7B model. 
To further verify the effectiveness of applying weight sharing to both MHA and FFN block, we should offer comprehensive experimental validation across different models and compare the results with baselines.  
We leave it as future work. 

Furthermore, the computing resources limited our ability to conduct experiments on LLMs with a model size of more than 13B. 
Although we hypothesize that our approach can still work in larger models, which proves to have redundant parameters~\citep{frantar2023sparsegpt}, it is crucial to validate this hypothesis with further exploration.

% \section*{Acknowledgements}

\bibliography{anthology,reference_new}
\bibliographystyle{acl_natbib}

\appendix
\input{7-Appendix}

\end{document}

%% file: 1-Introduction.tex
\section{Introduction}
Large Language Models (LLMs) have achieved breakthrough performance in a variety of natural language processing tasks ~\citep{wei2022emergent,bubeck2023sparks,zhao2023survey}. 
However, such remarkable capability typically comes at the cost of a substantial increase in the model size ~\citep{kaplan2020scaling}. 
Thus, LLMs with billions of parameters ~\citep{brown2020language,touvron2023llama} are more resource-hungry despite a wide margin of superiority over small-scale models~\citep{devlin2018bert,liu2019roberta}. 
This can also pose challenges for deployment on low-capability devices due to limited storage and GPU memory. 
% To address their high computational and memory requirements, several techniques are proposed to solve, like quantization ~\citep{bai2020binarybert,tao2022compression}, pruning ~\citep{yang2022gradient,tao2023structured} and knowledge distillation~\citep{wu2023ad,tan2023gkd}. 

To address the high memory requirements of models, weight sharing ~\citep{takase2021lessons,liu2023enhancing} aims to reuse the same parameters to achieve memory- and storage-efficiency while preserving model performance. 
For small-scale models, e.g., BERT, it is known that several techniques ~\citep{lan2019albert,liu2023enhancing} are proposed to explore across-layer parameter sharing. 
While, ~\citet{zhang2022minivit} demonstrate identical weights across different layers are the main cause of training instability and performance degradation. 
Moreover, the effectiveness of similar techniques at the scale of LLMs remains uncertain. 

Thus, we strive to solve this central question: \textbf{\textit{Can we design one fine-grained weight sharing strategy that can smoothly apply to large language models}?} 
For an effective memory-efficient weight sharing method tailored to LLMs, two key challenges must be tackled: a) the choice of shared modules whose weights are reused; b) the trade-off between reducing memory requirements and preserving diverse capabilities.

% our method
In the preliminary work, we empirically evaluate the feasibility of weight sharing across the attention heads in LLMs inspired by attention map (i.e., attention scores) reuse. 
Subsequently, we introduce our design of head-wise shareable attention strategy. 
It is a simple and intuitive technique for parameter sharing that can be implemented in a few minutes. 
Specifically, given the pre-trained weight matrices, we concatenate the weight matrix $W^q$ and $W^k$ for each head to measure the cosine similarity that determines which heads can be shared. 
Meanwhile, head-wise weight sharing promotes parameter diversity in the layers, and thus its performance degradation is acceptable when the number of shared parameters is below 30\%.
Even as weight sharing ratio increases rapidly, our proposed constrained post-training method can narrow the performance drop, which may necessitate additional time.

In summary, our key contributions include: 
\begin{itemize}[leftmargin=0.4cm,itemsep=0pt]
\item We investigate the feasibility of head-wise weight sharing for large language models and propose two corresponding methods named DirectShare and PostShare. 
% To our best knowledge, our proposal is the first work that explores the fine-grained weight sharing techniques for large PLMs. 
\item The proposed DirectShare is time-efficient and retain a large portion of the performance when sharing ratio is below 30\%. 
Complementarily, PostShare yields satisfactory performance via post-training, especially under large ratios. 
\item Experiments show our proposal achieves comparable performance to the competitive memory-efficient methods. 
Additional analysis also indicates its efficiency in small-scale models.
\end{itemize}

%% file: 2-Related_Works.tex
\section{Related Works}
\subsection{Memory-efficient Approaches for LLMs}
With the growing size of language models, several memory-efficient techniques are proposed to solve. 
One line to reducing the memory consumption involves network compression, like quantization ~\citep{bai2020binarybert,tao2022compression}, pruning ~\citep{yang2022gradient,tao2023structured,yang2024laco} and knowledge distillation~\citep{wu2023ad,tan2023gkd}. 
However, when applied to LLMs, many approaches have become infeasible ~\citep{frantar2023sparsegpt}. 
To recover performance, they require extensive post-training of the model ~\citep{dettmers2023spqr,sun2023simple}. 

In addition to these conventional methods, researchers have also investigated more efficient variations of the self-attention mechanism for LLMs ~\citep{kitaev2020reformer,lv2023lightformer}. 
Reformer ~\citep{kitaev2020reformer} leverages sparsity in the attention layers to improve the efficiency on long sequences and with small memory use. 
Lightformer ~\citep{lv2023lightformer} deploys SVD weight transfer and parameter sharing, which can significantly reduce the parameters on the premise of ensuring model performance. 
In this paper, our focus is on weight sharing across attention heads. 

\subsection{Weight Sharing}
Weight sharing is a widely used technique ~\citep{lan2019albert,liu2023enhancing,lv2023lightformer,xu2023compressionsurvey} that aims to improve parameter efficiency and reduce inference memory requirements. 
Weight sharing enables model compression by eliminating redundant parameters and decouples computation and parameters by reusing the same parameters for multiple computations.

\textbf{Task-oriented Weight Sharing.} 
One of the prevalent tasks using weight sharing mechanisms is nerual machine translation (NMT). Tied Transformer ~\citep{xia2019tied} considers model-level sharing and shares the weights of the encoder and decoder of an NMT model. ~\citet{dabre2019recurrent} proposes a method, which shares the weights across all Transformer layers and keeps performance in NMT. 
Besides, ~\citet{chi2021audio} bring the idea of ALBERT ~\citep{lan2019albert} to the speech recognition task. 

\textbf{Layer-wise Weight Sharing.}
Universal Transformer ~\citep{dehghani2018universal} shares the weights across all layers with a dynamic halting mechanism and improves accuracy on several tasks. 
Subformer ~\citep{reid2021subformer} utilizes sandwich-style parameter sharing, which only shares the central layers while leaving the first and last layers independent. 
~\citet{takase2021lessons} study strategies to explore the best way to prepare parameters of M layers and assign them into N layers (1$\leq$M$\leq$N). 

%% file: 3-Motivation.tex
\section{Motivation and Empirical Analysis}\label{sec:motivation}
In this section, we analyze the feasibility of head-wise weight sharing from the perspective of attention map reuse. 
% Section~\ref{sec:attention_map} delves into % We start by visualizing 
% attention map similarity, specifically transitioning from layer-wise to head-wise analysis. 
% Section~\ref{sec:weight_similar} investigates a potential relationship between attention map similarity and weight similarity. 

\begin{figure*}[!tb]
    \centering
    \includegraphics[width=\textwidth]{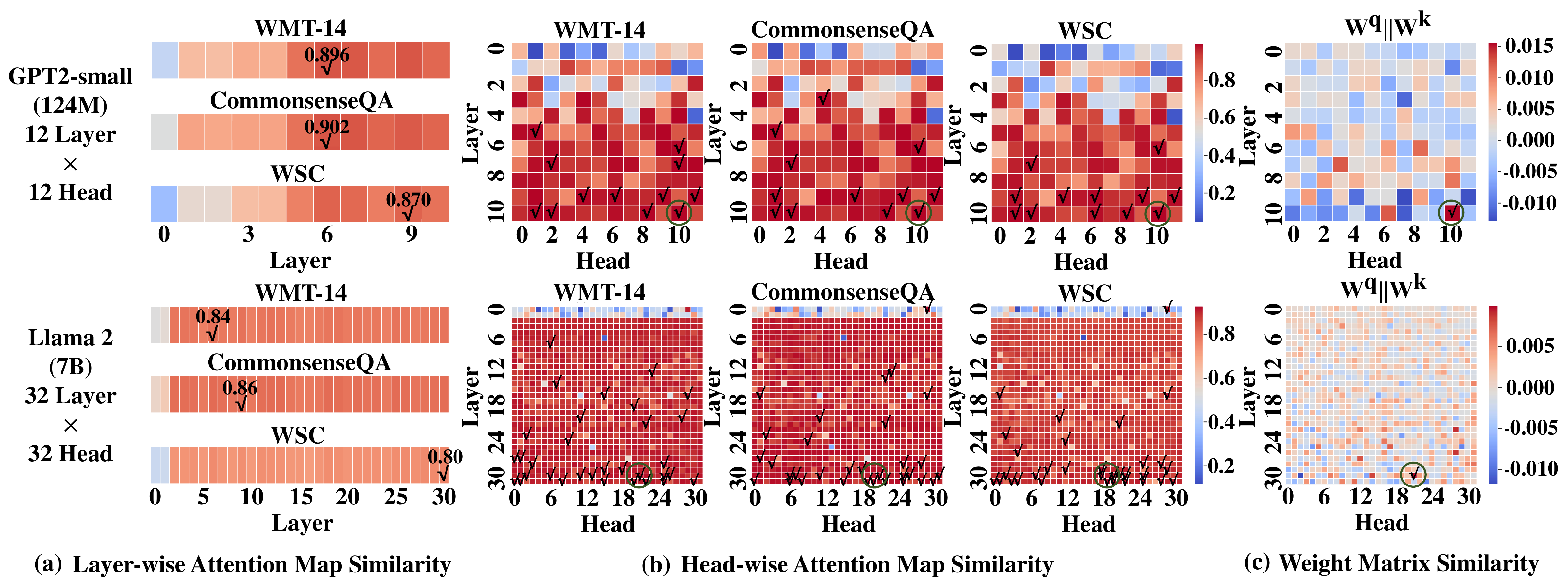}
    \caption{(a) Layer-wise Attention Map Similarity. Taking the last layer as an example, the most similar attention layer with it is marked with $\surd$. (b) Head-wise Attention Map Similarity. $\surd$ mark the top n heads whose attention maps that are most similar to the 6-th head in the last layer(n=the number of heads per layer). (c) Weight Matrix Similarity. \color{green!50!black}{\textbf{$\bigcirc$}}\color{black}{ mark the connection between attention map similarity and weight similarity.}}
    \label{fig:similarity}
\end{figure*}

\subsection{Attention Map Similarity: From Layer-wise to Head-wise}\label{sec:attention_map}
Prior researches \citep{xiao2019sharing,ying2021lazyformer,bhojanapalli2021leveraging} demonstrate the effectiveness of attention map reuse due to the high similarity of attention scores between different layers (especially for adjacency layers).
Motivated by this, we delve into attention map similarity, specifically transitioning from layer-wise to head-wise analysis. 
To measure the evolution of the attention maps over layers and heads, we use the cosine similarity $\mathcal{S}_{cos}$. 
When $\mathcal{S}_{cos}$ equals one, it means that the attention maps are perfectly similar. 
Considering two specific self-attention layers, the cosine similarity is calculated as follows:
\begin{equation}
    \mathcal{S}_{cos}(\textbf{A}_p,\textbf{A}_q)=\frac{\textbf{A}_p^T\textbf{A}_q}{\|\textbf{A}_p\| \|\textbf{A}_q\|}
\end{equation}
where $\textbf{A}_p,\textbf{A}_q$ denote the attention map of layers p and q.

We visualize the layer-wise and head-wise attention map similarity across three task-specific datasets: WMT14 (En-Fr) \citep{bojar-EtAl:2014:W14-33}, CommonsenceQA \citep{talmor-etal-2019-commonsenseqa} and WSC \citep{levesque2012winograd}. 
As shown in Fig.~\ref{fig:similarity}(a) and (b), the degrees of similarity in attention scores computed in different layers and heads present a certain level of consistency across different tasks.
In addition, we find that the cosine similarity values for pairs with high similarity are higher among different heads compared to layers.  
Specifically, the most similar self-attention layers reach a cosine similarity value of approximately 0.90, while in the case of head-wise comparisons, several pairs have a  remarkable similarity of nearly 0.99. 

One observation is that as the number of parameters increases, modules with high similarity exhibit variations, particularly in the fine-grained (e.g., head-wise) comparisons within large-scale pre-trained language models. 
Existing approaches employ "learning to share" techniques to dynamically adjust the sharing strategy~\citep{xiao2019sharing} or use a uniform sharing strategy but train the modified model from scratch~\citep{ying2021lazyformer,shim2023exploring}. 
However, such strategies pay little attention to reusing attention map among heads and incur high computational costs for LLMs. 

\vspace{-5pt}
\subsection{From Attention Map Similarity to Weight Matrix Similarity}\label{sec:weight_similar}
Attention weight matrix similarity provides a complementary perspective to attention map similarity, since the attention scores are calculated based on the weight matrices $W^q,W^k$. 
Weight sharing is traditionally based on the assumption that overparameterization is evident in large-scale Transformer models, i.e., the difference in weights decreases as model size increases~\citep{li2020train}. 
In this paper, we explore a potential relationship between attention map similarity and weight similarity.

As mentioned in Section~\ref{sec:attention_map}, head-wise attention map similarity is higher than the cross-layer similarity, while to the best of our knowledge, head-wise attention map reuse is yet to be explored. 
This might be attributed to the difficulty in finding an optimal dynamic head-wise sharing strategy across different tasks. 
One intuitive solution is to first measure the attention map similarity between every pair of heads in each dataset separately, and then choose the overlapping modules to share. 

Combined with the analysis of weight matrix similarity, we have made a key discovery: given a pre-trained LLM, by concatenating the weight matrix $W^q$ and $W^k$ for each head to measure the cosine similarity, the most similar weight matrix corresponds to the overlapping modules with highly similar attention maps observed across different datasets. 
As illustrated in Fig.~\ref{fig:similarity}(b) and (c), deep green circles mark the connection between attention map similarity and weight similarity (more analysis in Appendix~\ref{app:similarity}).

This finding implies that attention heads with high weight matrix similarity also demonstrate analogous attention map similarity regardless of the datasets and model size. 
Furthermore, since different heads within the layer present sufficient diversity~\citep{zhou2021deepvit,vig2019multiscale}, we suppose that weight sharing among these heads can result in higher model behavior consistency compared to layer-wise weight sharing. 
Thus, we further propose a simple yet effective method for head-wise weight sharing, especially validating its feasibility in large-scale models. 

%% file: 4-Method.tex
\begin{figure*}[htbp]
    \centering
    \includegraphics[width=\textwidth]{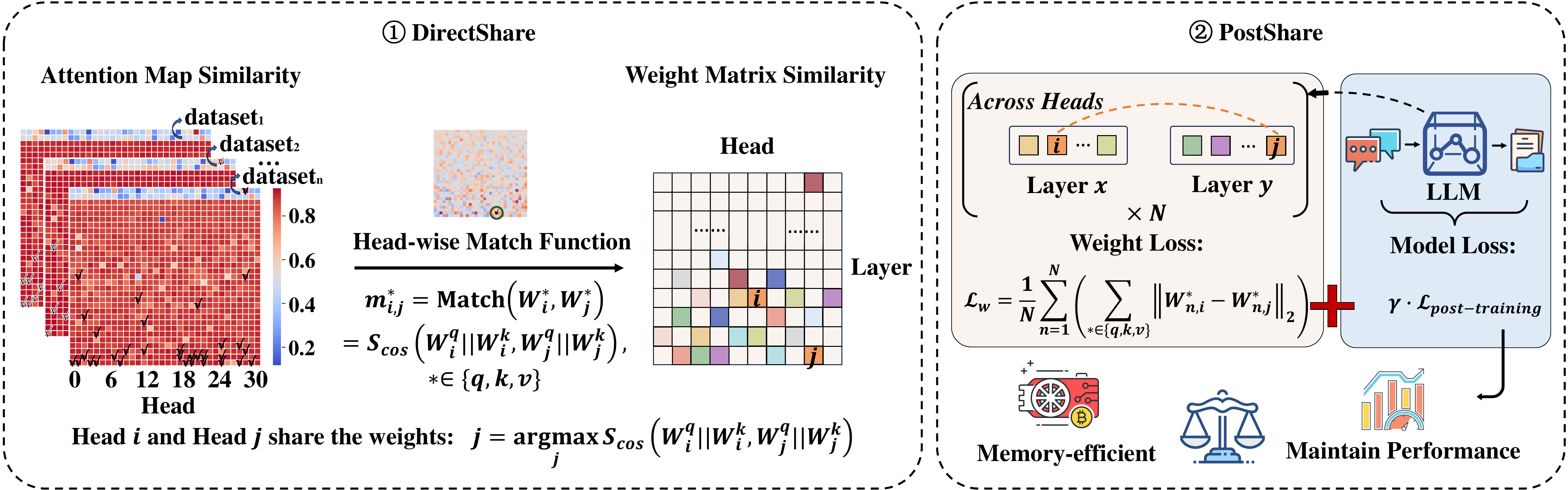}
    \caption{% Components of our proposed Head-wise Shareable Attention. 
    \ding{172} \textbf{DirectShare}: Inspired by attention map reuse, directly share weight matrices across different heads based on cosine similarity; \ding{173} \textbf{PostShare}: To balance the memory usage and the performance, implement post-training with the constraint of weight matrix similarity and then share.}
    \label{fig:pipeline}
\end{figure*}

\section{Head-wise Shareable Attention}\label{sec:method}
Inspired by Section~\ref{sec:motivation}, we present a perspective on head-wise shareable attention for LLMs. 
Based on one straightforward yet effective weight sharing strategy, we propose two complementary methods, named \textbf{DirectShare} and \textbf{PostShare}. 
The overview of our proposal is presented in Figure~\ref{fig:pipeline}. 

\subsection{Head-wise Weight Sharing Strategy}\label{sec:strategy}
Multi-Head Attention (MHA) block is essentially a procedure  that computes the relevance of each token in a sentence with respect to all other tokens. 
Let $L$ be the number of input tokens and $M$ be the number of attention heads in total. 
Given the input $X\in \mathbb{R}^{L\times D}$, we can obtain queries, keys, and values in the $i$-th ($1$$\leq$$i$$\leq$$M$) head via three weight matrices, denoted by $W^q_i \in \mathbb{R}^{D\times d_q}$, $W^k_i \in \mathbb{R}^{D\times d_k}$ and $W^v_i \in \mathbb{R}^{D\times d_v}$, respectively. 
$D$ is the embedding dimension, and $d_q, d_k (=d_q), d_v$ represent the dimensions of three weight matrices, respectively. 

\begin{figure}[!t]
    \centering
    \includegraphics[width=\linewidth]{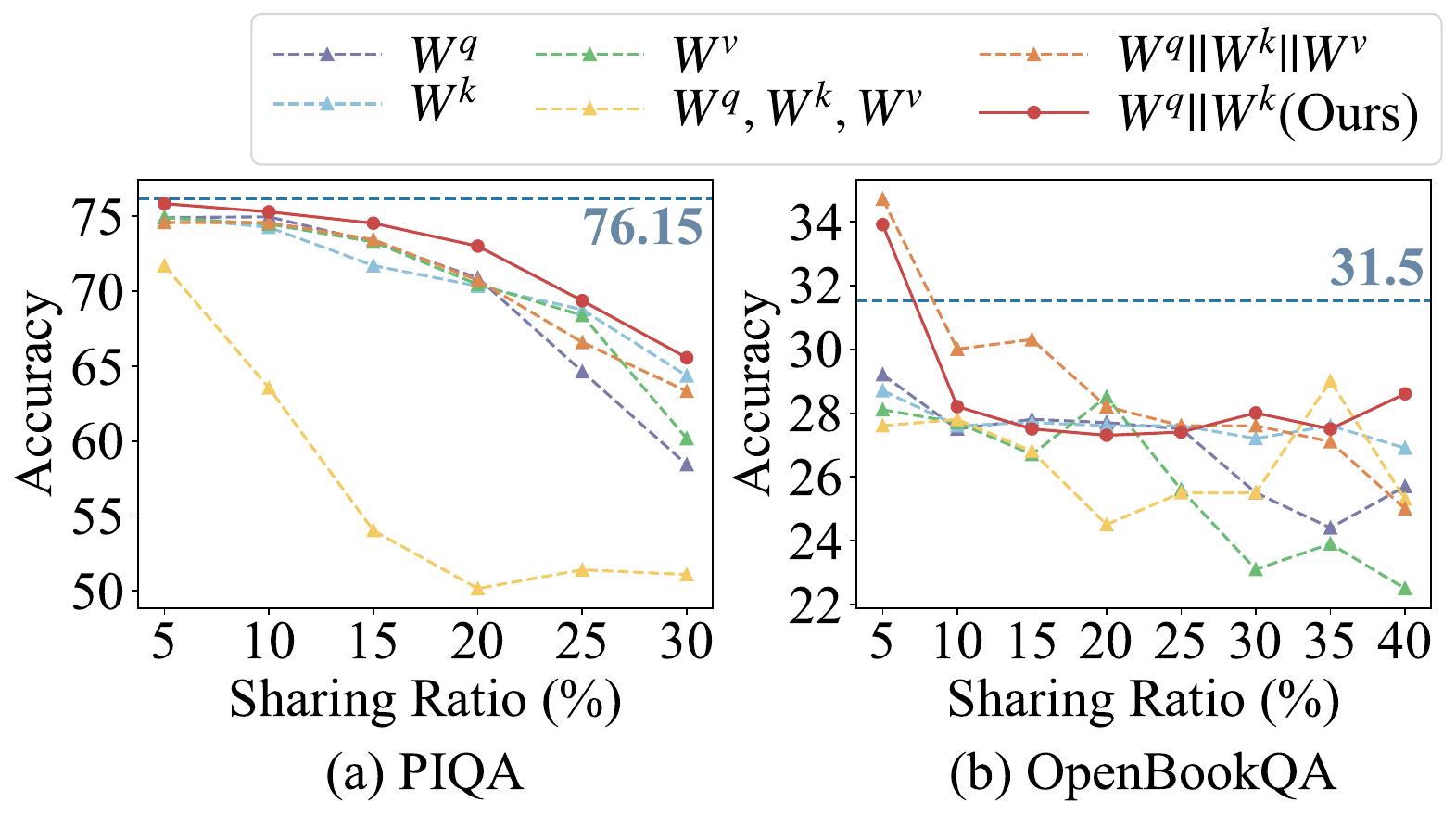}
    \setlength{\abovecaptionskip}{-10pt}
    \setlength{\belowcaptionskip}{-10pt}
    \caption{Experiments performed on PIQA and OpenBookQA using different head-wise match functions for Baichuan2-7B model.}
    \label{fig:sharing_strategy}
\end{figure}

To investigate the strategy of weight sharing applied to all the above three weight matrices across heads for LLMs, we perform preliminary experiments in the choice of head-wise match functions \textbf{Match(}$\cdot$,$\cdot$\textbf{)}.  
For the match functions, inputs are the weight matrices of head $\bm{i,j}$ and outputs are called matching scores $\bm{m}$. 
The higher the score, the more likely it is to share parameters across the heads. 
\begin{equation}
    \bm{m_{i,j}^*}=\textbf{Match(}W_i^*,W_j^*\textbf{)},* \subseteq \{q,k,v\}
\end{equation}
Based on our intuitive analysis in Section~\ref{sec:weight_similar}, we choose the cosine similarity between the concatenation matrix of $W^q_i$ and $W^k_i$: 
\begin{equation}
% \resizebox{0.85\linewidth}{!}{$
    \bm{m_{i,j}^q}=\bm{m_{i,j}^k}=\bm{m_{i,j}^v}=\mathcal{S}_{cos}(W^q_i||W^k_i,W^q_j||W^k_j)
% $}
\label{eq:match_func}
\end{equation}
Besides, we try another five match functions to compare: (1) Only $W^q_i$ used to measure the cosine similarity, i.e., $\bm{m_{i,j}^*}=\mathcal{S}_{cos}(W^q_i,W^q_j)$; (2) Only $W^k_i$ used to measure the cosine similarity, i.e., $\bm{m_{i,j}^*}=\mathcal{S}_{cos}(W^k_i,W^k_j)$; (3) Only $W^v_i$ used to measure the cosine similarity, i.e., $\bm{m_{i,j}^*}=\mathcal{S}_{cos}(W^v_i,W^v_j)$; (4) Concatenate all the three matrices and then calculate the cosine similarity, i.e., $\bm{m_{i,j}^*}=\mathcal{S}_{cos}(W^q_i||W^k_i||W^v_i,W^q_j||W^k_j||W^v_j)$; (5) Separately use $W^q_i, W^k_i, W^v_i$ to measure the cosine similarity and do weight sharing respectively, i.e., $\bm{m_{i,j}^*}=\mathcal{S}_{cos}(W^*_i,W^*_j)$ and again $*\in\{q,k,v\}$.

Figure~\ref{fig:sharing_strategy} shows the results of our exploratory study via DirectShare. 
As evidenced by the performance curve, using separately weight sharing causes a significant decline in performance compared with sharing the three weight matrices together.
And it is enough to do head-wise weight sharing focusing only on the concatenation matrix of $W^q_i$ and $W^k_i$, since it achieves a favorable trade-off between reducing memory footprint and maintaining performance. 

\subsection{DirectShare}
In practice, we traverse all head pairs to compute matching scores on Equation~\ref{eq:match_func} and for each head, select the one with the highest score to match. 
When candidate shareable head pairs prepared, we select the top-N pairs in descending order according to the desired sharing ratio $\alpha$. 
Finally, we can share the weight matrices together
between each selected attention head pairs. 
A detailed algorithm for our DirectShare is presented in Algorithm~\ref{alg:sharing} and Appendix~\ref{app:alg}.
\begin{algorithm2e}
% \SetAlgoLined % for end
\caption{\normalsize {%Workflow of 
\textbf{DirectShare} using Head-wise Weight Sharing Strategy
}}\label{alg:sharing}
\SetInd{0.5em}{0.6em}
\KwIn{Sharing ratio $\alpha$, Original LLM $\mathcal{M}$, \\Number of layers $\mathcal{L}$, \\Number of heads per MHA block $\mathcal{H}$}
\KwOut{The LLM $\mathcal{M^{*}}$ after weight sharing}
Initialize candidate buffer $\mathcal{D}_\tau$\;
\For{$layer_i\leftarrow 2$ \KwTo $\mathcal{L}$}{
    \For{$i\leftarrow 1$ \KwTo $\mathcal{H}$}{
        $index_i \leftarrow$ $(layer_i,i)$\;
        $index_m$ $\leftarrow$ None\;
        $s_m$ $\leftarrow$ -$1$\;
        \For{$layer_j\leftarrow 1$ \KwTo $layer_i-1$}{
            \For{$j\leftarrow 1$ \KwTo $\mathcal{H}$}{
                $index_m \leftarrow (layer_j,j)$\;
                Compute $\bm{\mathcal{S}_{cos}}$ using Eq.~\ref{eq:match_func}\;
                \If{$\bm{\mathcal{S}_{cos}}>s_m$}{
                    $s_m \leftarrow \bm{\mathcal{S}_{cos}}$ 
                }
            }
        }
        Store candidate shareable head pair $<index_i,index_m,s_m>$ in $\mathcal{D}_\tau$\;
    }
} 
Sort $\mathcal{D}_\tau$ by descending matching scores $s_m$\;
$\mathcal{N}$ $\leftarrow$ Top\_N($\mathcal{D}_\tau$, $\mathcal{L}$, $\mathcal{H}$, $\alpha$)\; 
$\mathcal{M^{*}}$ $\leftarrow$ Weight\_Share($\mathcal{M}$, $\mathcal{N}$).
\end{algorithm2e}

\subsection{PostShare}% Fast Post-training Recovery
Although DirectShare demonstrates effectiveness in our experiments, we have also encountered noticeable performance drop in minor reading comprehension datasets. 
To alleviate this problem, we propose PostShare, softly aligning model weights during the post-training process. 

With the same sharing strategy (Section \ref{sec:strategy}), PostShare first selects the set of weight matrices to share. 
Next, we incorporate a regularization term into the loss function to constrain our post-training process, encouraging selected weight matrices more similar: 
\begin{equation}
    \mathcal{L}_w=\frac{1}{|\mathcal{N}|}\sum_{(i,j)\in \mathcal{N}}\left( \sum_{* \in \{q,k,v\}} \left\| W_{i}^*-W_{j}^*\right\|_2\right)
\end{equation}
where $\mathcal{N}$ is the set of selected attention head pairs for sharing. 
With this regularization weight loss, the proposed PostShare learn model weights by minimizing the following combined loss function: 
\begin{equation}
    \mathcal{L}=\mathcal{L}_{post-training}+\gamma \times \mathcal{L}_w
\end{equation}
where $\mathcal{L}_{post-training}$ is the original post-training loss, $\gamma$ controls the strength of $\mathcal{L}_w$. 
After the post-training process, the corresponding weight matrices can be shared as DirectShare does. 
Although post-training indeed increases the time cost of weight sharing, PostShare achieves stable and satisfactory performance across different tasks when reducing memory usage. 
% push selected weights closer

%% file: 5-Experiments.tex
\section{Experiments}
\begin{table*}[htbp]
\centering
\setlength\tabcolsep{3pt} 
\renewcommand\arraystretch{1.6}
\Huge
\resizebox{\linewidth}{!}{
\begin{tabular}{@{}m{2cm}m{5cm}m{2.5cm}m{2.5cm}m{2.4cm}m{2.4cm}m{2.4cm}ccccccccc@{}}
\toprule
\toprule
\multicolumn{2}{l}{\textbf{Benchmark Type}}&\multicolumn{5}{c}{\textbf{Reasoning}}&\multicolumn{5}{c}{\textbf{NLU}}& \multicolumn{4}{c}{\textbf{Knowledge}}\\
\cmidrule(lr){3-7}\cmidrule(lr){8-12}\cmidrule(lr){13-16}
\midrule
\textbf{Ratio} & \textbf{Method}
& \hspace{-13pt}CMNLI & OCNLI & \hspace{5pt}AX-b & AX-g & RTE & \makecell{ RACE- \\ middle} & \makecell{ RACE- \\ high} & 
OBQA %& \textbf{\makecell{OBQA- \\ fact}} 
& \hspace{5pt}CSL & TNEWS & \makecell{Wino-\\Grande}& BoolQ & C-Eval & MMLU
\\
\midrule
\textbf{0\%} & \textbf{Llama2-7B} &  32.98 &33.12 &53.53&55.34&49.82&  33.15 &35.51&31.80&55.62&20.22&  54.04&70.67&32.20&46.69\\
 \midrule
\multirow{3}{*}{\textbf{10\%}} &Magnitude&\underline{32.99}&30.63&\underline{56.70}&49.44&47.29&25.42&26.47&\textbf{28.20}&49.38&14.85&51.58&60.80&22.16&28.20\\
& LLM-Pruner & \underline{32.99}& \textbf{33.75}&\textbf{57.61}&\underline{50.00}&\underline{48.38} & \underline{28.20}&\textbf{30.73}&\underline{27.20}&\underline{53.12}&\underline{19.76}& \textbf{52.98}&\underline{66.09}&\underline{22.31}&\underline{38.11} \\
& DirectShare  &\textbf{33.00}&\underline{32.50}&54.17&\textbf{51.97}&\textbf{50.90}&\textbf{28.34}&\underline{28.96}&\textbf{28.20}&\textbf{54.37}&\textbf{20.86}&\underline{52.63}&\textbf{67.74}&\textbf{28.75}&\textbf{43.43}\\
 \midrule
\multirow{3}{*}{\textbf{30\%}}
&Magnitude&\underline{33.16}&\textbf{35.00}&54.71&50.56&46.93&\textbf{21.80}&\underline{21.53}&25.00&45.62&7.01&\textbf{50.88}&44.59&\underline{24.38}&23.15\\
& LLM-Pruner& 32.99&31.25 &\underline{56.34}&\textbf{52.53}&\underline{48.74}& \underline{21.52}&\textbf{22.21} &\textbf{26.80}&\underline{50.00}&\underline{10.20}& \textbf{50.88}&\textbf{54.77}&22.82&\underline{25.16}\\
& DirectShare  & \textbf{33.33} &\underline{32.50}&\textbf{57.07}&\underline{51.69}&\textbf{49.10}& 21.45&\underline{21.53}&\underline{26.00}&\textbf{51.25}&\textbf{20.22}& \underline{50.18}&\underline{54.43} &\textbf{26.24}&\textbf{26.53}\\ 
\midrule
\midrule
\textbf{0\%} & \textbf{Llama2-13B} &32.99&35.00&58.81&50.56&47.29&60.24 & 58.03&42.40&58.75&22.13&55.44&71.50&40.17&55.81\\
 \midrule
\multirow{3}{*}{\textbf{10\%}}
&Magnitude&\underline{32.82}&\underline{33.12}&51.99&\textbf{50.56}&\textbf{48.38}&22.42&21.78&27.40&51.25&15.39&49.82&62.32&22.52&27.54\\
& LLM-Pruner &\textbf{32.99}&\textbf{36.25}&\textbf{58.70}&\underline{50.00}&46.93 & \underline{51.46}&\underline{50.80}&\textbf{47.00}&\underline{56.25}& \textbf{20.95}& \textbf{55.44}&\underline{68.07}&\underline{30.25}&\underline{51.45}\\
& DirectShare  & \textbf{32.99}&\textbf{36.25}&\underline{57.61}&\underline{50.00}&\underline{47.29}& \textbf{54.04}&\textbf{55.63}&\underline{39.40}&\textbf{56.88}&\underline{17.94}& \underline{54.39}&\textbf{69.45}&\textbf{37.17}&\textbf{52.81}\\
 \midrule
\multirow{3}{*}{\textbf{30\%}}
&Magnitude&\textbf{33.78}&33.75&46.65&\underline{50.00}&\textbf{51.99}&21.80&22.01&\textbf{28.80}&46.25&4.19&49.12&56.45&\textbf{23.99}&22.86\\
& LLM-Pruner &\underline{32.99}&\underline{34.38}&\underline{57.16}&\textbf{54.21}&45.85 & \underline{23.96}&\underline{25.33}&26.40&\underline{53.75}&\textbf{16.76}& \textbf{51.58}&\textbf{63.21}&22.17&\underline{27.22}\\
& DirectShare &\underline{32.99}&\textbf{35.00}&\textbf{58.33}&\underline{50.00}&\underline{46.57}&\textbf{26.53}&\textbf{27.53}&\underline{27.40}&\textbf{59.38}&\underline{16.12}&\underline{50.18}&\underline{59.36}&\underline{22.30}&\textbf{30.79}\\ 
\bottomrule
\bottomrule
\end{tabular}
}
\setlength{\abovecaptionskip}{5pt}
\setlength{\belowcaptionskip}{-5pt}
\caption{Evaluation results of DirectShare based on the Llama2-7B and Llama2-13B models. \textbf{Bold} and \underline{underline} indicate the best and the second best results.}
\label{tab:overall_results}
\end{table*}

\subsection{Experimental Settings}
\textbf{Models.} 
We evaluate DirectShare and PostShare on two open-source LLMs: Llama2~\citep{touvron2023llama} and Baichuan2~\citep{baichuan2023baichuan2} with 7B and 13B parameters. 
In PostShare, we use English Wikipedia~\citep{wikidump} to post-train the backbone models for weight sharing. 

\noindent
\textbf{Evaluation.} 
To comprehensively evaluate the model capabilities, we experiment on five distinct tasks: reasoning, understanding, language, knowledge and examination. 
For consistent comparisons, we deploy open-source LLM evaluation platform OpenCompass~\citep{2023opencompass}. 

%\textbf{Implementation Details.} 
\noindent
\textbf{Baselines.} 
% To our best knowledge, our proposal is the first work that explores the fine-grained weight sharing techniques for large PLMs. 
Since existing weight sharing techniques do not support LLMs, %we need to re-implement them to accommodate. 
% In this study, we select \textbf{CYCLE(REV)}~\citep{takase2021lessons} designed for Transformers, which applies layer-wise weight sharing. 
% Alongside such weight sharing methods, we also 
we compare DirectShare against Magnitude Pruning~\citep{zhu2017prune} and LLM-Pruner~\citep{ma2023llm}, two influential works for model pruning. 
Certainly, they are not directly comparable. 
To ensure fairness in the experiments, both of them only prune the multi-head attention module and thus compare when the same number of parameters is reduced. 
See Appendix~\ref{app:experiments} for additional information. 
% With the aim of memory reduction, we generalize Wanda to structured pruning following~\citet{an2023fluctuation} re-implementation called Wanda-sp.   

\subsection{Main Results} 
\subsubsection{Evaluation on DirectShare}\label{sec:directshare}

Table~\ref{tab:overall_results} shows the overall performance of DirectShare based on Llama2 models. 
Benchmarks are classified into three categories: reasoning, natural language understanding (NLU) and knowledge-related. 
The corresponding results for Baichuan2 models can be found in Appendix~\ref{sec:baichuan2}. 

\noindent
\textbf{Logical and Common Sense Reasoning.} 
In the domain of reasoning, when applying a 30\% parameter sharing to Llama2-7B, our DirectShare can still maintain an average performance of 99.51\% across the five benchmarks, compared to the base model. 
With the same setting, the shared Llama2-13B retains 99.21\% performance. 
This suggests our finding of head-wise shareable attention for LLMs indeed can work without significant performance degradation in reasoning tasks. 

The overall efficacy of our DirectShare rivals with the structured pruning results of LLM-Pruner, without any training. 
Moreover, our method is quite simple and fast, independent on the original training corpus, while structured pruning will nearly fail in the zero-shot generation tasks without dependencies~\citep{ma2023llm}.
%even surpassing the performance of the original model
%indicating its strong capabilities 

\noindent
\textbf{Natural Language Understanding (NLU).} 
Compared to reasoning tasks, our experimental results unveil a notable performance decrease of approximately 30\% in large-scale reading comprehension datasets when applying  DirectShare to Llama2-7B model. 
Beyond this, we discover that on content summary and analysis tasks, DirectShare manages to retain 94.23\% of the performance exhibited by the base model. 
The evaluation results of Llama2-13B align with those of Llama2-7B and we find the accuracy gap is larger as model size increases. 
This trend also exists in Magnitude Pruning and LLM-Pruner, even the performance drop is larger: LLM-Pruner drops $\approx$ 3 points more than ours on average while Magnitude Pruning is outperformed by ours by a large margin. 

To mitigate this degradation, some post-training pruning methods like SparseGPT~\citep{frantar2023sparsegpt} preserves accuracy via the weight update procedure. 
Similarly, LLM-Pruner uses the low-rank approximation (LoRA, ~\citealp{hu2021lora}) to post-train the pruned model. 
% This makes it possible to merge regularization weight loss into a post-training procedure. 
Motivated by this, our PostShare proves to be beneficial, substantially improving 17.80\% accuracy, albeit at a certain amount of time cost. 
For more details refer to Section~\ref{sec:postshare}. 
However, this does not diminish the significance of our DirectShare. 
The absence of post-training allows us to better understand the feasibility of head-wise weight sharing for LLMs.

\begin{table*}[t]
\centering
\setlength\tabcolsep{3pt} 
\large
\resizebox{\textwidth}{!}{
\begin{tabular}{@{}cccccccccccc@{}}
\toprule
\toprule
\textbf{Ratio} & \textbf{Method}
& \textbf{WinoGrande}& \textbf{BoolQ} & \textbf{C-Eval} & \textbf{MMLU}&  \textbf{RACE-middle} & \textbf{RACE-high} & \textbf{OBQA}& \textbf{OBQA-fact}
\\
\midrule
\textbf{0\%} & \textbf{Llama2-7B} & 54.04&70.67 & 32.20 & 46.69 & 33.15 & 35.51 & 31.8&42.2\\
\midrule
\multirow{2}{*}{\textbf{30\%}}& DirectShare & 50.18&54.43 & 26.24 & 26.53 &  21.45 & 21.53& 26.00&27.60\\
& PostShare &52.98 \small{\textcolor{red}{$\uparrow2.80$}}&66.57 \small{\textcolor{red}{$\uparrow12.14$}}&26.38 \small{\textcolor{red}{$\uparrow0.14$}}&33.36 \small{\textcolor{red}{$\uparrow6.83$}}&29.81 \small{\textcolor{red}{$\uparrow8.36$}}&29.45 \small{\textcolor{red}{$\uparrow7.92$}}&27.60 \small{\textcolor{red}{$\uparrow1.60$}}&33.60 \small{\textcolor{red}{$\uparrow6.00$}}\\ 
\bottomrule
\bottomrule
\end{tabular}
}
\setlength{\belowcaptionskip}{-10pt}
\caption{Overall Performance of PostShare based on Llama2-7B model. See Appendix~\ref{app:postshare_13b} for results on Llama2-13B.}
\label{tab:postshare}
\end{table*}

\begin{figure}[htbp]
    \centering
    \includegraphics[width=\linewidth]{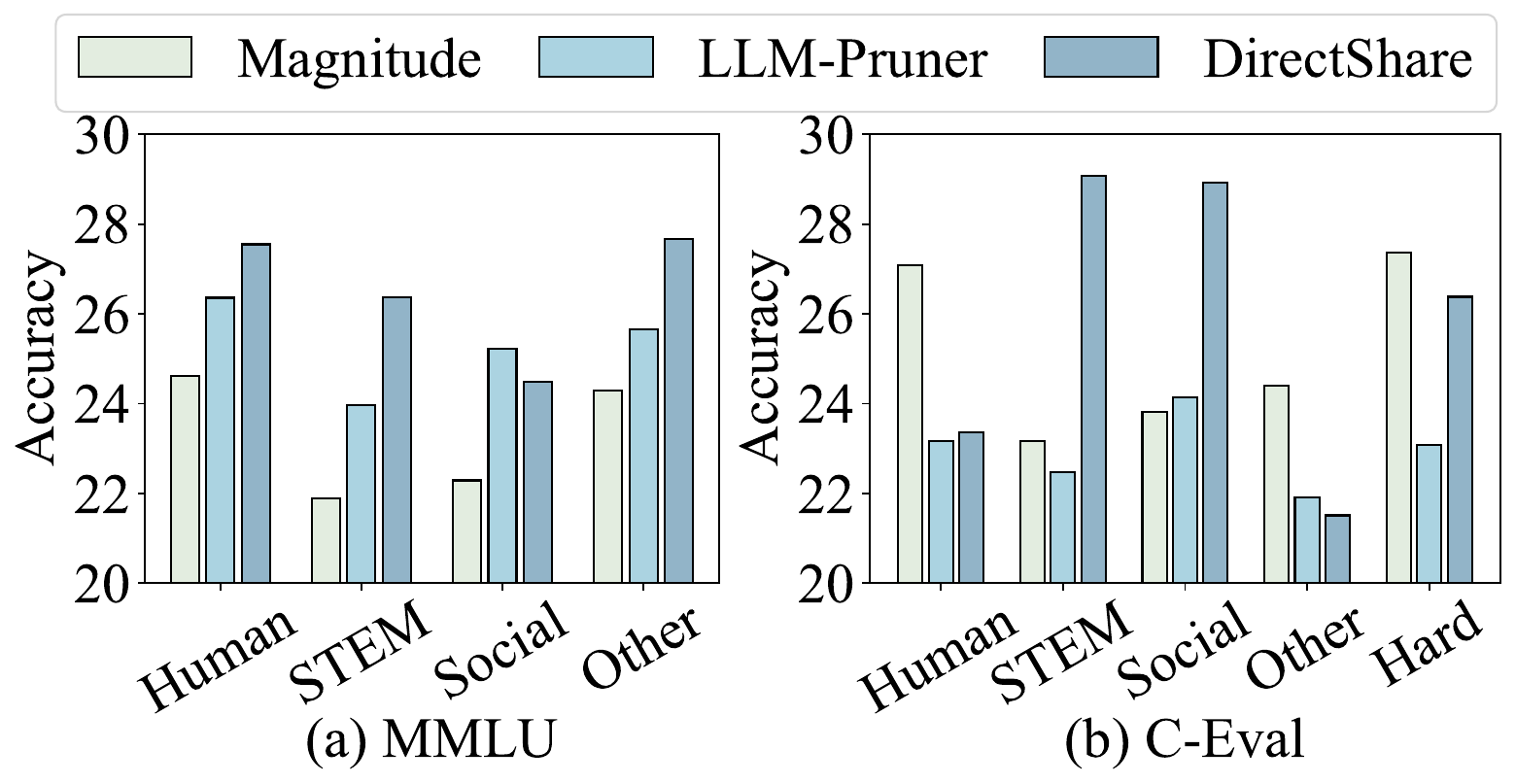}
    \setlength{\abovecaptionskip}{-10pt}
    \setlength{\belowcaptionskip}{-15pt}
    \caption{Performance of DirectShare across different subjects based on Llama2-7B on C-Eval and MMLU. %Human stands for humanities; Social stands for social science.
    }
    \label{fig:exam}
\end{figure}

\noindent
\textbf{Knowledge-related Tasks.} 
As depicted in Table~\ref{tab:overall_results}, DirectShare takes a clear advantage over other approaches in the field of examination. 
Our chosen C-Eval and MMLU span diverse disciplines to test both world knowledge and problem solving ability exclusively in a Chinese and English context, respectively. 
To make this more concrete, Figure~\ref{fig:exam} vividly contrasts the performance across different subjects based on Llama2-7B on C-Eval and MMLU. 
But we have to admit directly do weight sharing across attention heads results in a obvious decline in knowledge-related abilities, which can be solved in PostShare. 

\subsubsection{Evaluation on PostShare}\label{sec:postshare}
Based on the evaluation conducted on DirectShare, we experiment on PostShare, with a special focus on those benchmarks where DirectShare experiences a large accuracy degradation. 

Table~\ref{tab:postshare} reports how the performance improves with only 0.5 training epoch for Llama2-7B model. 
Specifically, in the reading comprehension and knowledge-related tasks mentioned above, PostShare achieves 87.53\% of the overall accuracy attained by the original model. 
Most of the gap between models after DirectShare and the original counterparts can be narrowed via PostShare, especially in BoolQ and RACE datasets. 

Last, it is important to emphasize that here we perform post-training with limited training corpus and thus it runs the risk of overfitting when training only for one epoch. 
For example, PostShare achieves the higher accuracy in BoolQ at 0.3 epoch than at 0.5 epoch (\underline{68.29} vs. 66.57). 
In contrast, as the training epoch increases from 0.5 to 0.9, the accuracy in WinoGrande rises (52.98 vs. \underline{54.39}). 
It means that due to the domain-constrained corpus, 
overfitting to one specific dataset will potentially compromise the capabilities in other tasks. 
The in-depth analysis is provided in Appendix~\ref{app:overfitting}. 

\subsection{Additional Analysis}\label{sec:addition_study}

\textbf{Statistics of Memory Reduction.} 
Table~\ref{tab:memory} presents the statistics of the parameter count and memory requirements when applying DirectShare. 
When sharing 30\% parameter sharing in the MHA block, our method achieves 10-13\% memory. 

Moreover, we find our weight sharing strategy (Section~\ref{sec:strategy}) also applies to FFN block. 
We directly observe the weight matrix similarity in FFN and find the concatenation matrix of gate\_proj, up\_proj and down\_proj can be used as matching function for FFN block. 
Since FFN does not have explainable fine-grained sub-blocks (like attention heads in MHA), we use Traversal Searching method to choose the optimal size of sub-block and find sharing the whole FNN layer works best in the performance maintenance.
Finally, when we share 30\% of parameter sharing in both MHA and FFN block, the model can save 26-28\% GPU memory. 
\begin{table}[htbp]
    \renewcommand\arraystretch{1.15}
    \centering
    \normalsize
    \resizebox{0.95\linewidth}{!}{
    \begin{tabular}{lll}
    \toprule[0.2pt]
    \toprule[0.2pt]
    \textbf{Sharing Ratio} &\textbf{\#Params} & \textbf{GPU Memory}\\
    \midrule
   \rowcolor{gray!20} 
\multicolumn{3}{c}{\textbf{Llama2-7B}}\\
    0\%	&6.74B/100\%	&17826M/100\%	\\
    30\% MHA	&6.09B/90.36\%	&15512M/87.02\%	\\
    30\% MHA+FFN &4.74B/70.33\%	&12932M/72.55\%	\\
    \rowcolor{gray!20} 
\multicolumn{3}{c}{\textbf{Llama2-13B}}\\
    0\%	&13.02B/100\%&30800M/100\%	\\
    30\% MHA &	11.76B/90.32\%	&27898M/90.58\%	\\
    30\% MHA+FFN &9.21B/70.74\%&23002M/74.68\%\\
    \bottomrule[0.2pt]
    \bottomrule[0.2pt]
    \end{tabular}}
    \caption{The actual memory savings brought by DirectShare on Llama2 models (recorded during inference on the BoolQ Dataset in OpenCompassv1.0 platform).}
    \label{tab:memory}
\end{table}

\begin{table*}[t]
\centering
\vspace{-15pt}
\setlength\tabcolsep{3pt} 
\large
\resizebox{\textwidth}{!}{
\begin{tabular}{@{}ccccccccccccccccc@{}}
\toprule
\toprule
\textbf{\makecell{Method\\Ratio=30\%}} & \textbf{CMNLI}& \textbf{OCNLI}& \textbf{AX-b}& \textbf{AX-g}& \textbf{RTE} & \textbf{\makecell{Wino-\\Grande}}& \textbf{BoolQ} & \textbf{C-Eval} & \textbf{MMLU}&  \textbf{\makecell{RACE-\\middle}} & \textbf{\makecell{RACE-\\high}} & \textbf{OBQA}& \textbf{\makecell{OBQA-\\fact}}&\textbf{CSL}
\\
\midrule
%  \textbf{Llama 2-7B} && 54.04&70.67 & 32.20 & 46.69 & 33.15 & 35.51 & 31.8&42.2\\
% \midrule
DirectShare &33.33&32.50&57.07&51.69&49.10& 50.18&54.43 & 26.24 & 26.53 &  21.45 & 21.53& 26.00&27.60&51.25\\
\midrule
\makecell{DirectShare \\+ 4bit GPTQ} &\makecell{34.61\\\small{\textcolor{red!75!black}{$\uparrow1.28$}}}&\makecell{30.63\\\small{\textcolor{green!50!black}{$\downarrow1.87$}}}&\makecell{57.79\\\small{\textcolor{red!75!black}{$\uparrow0.72$}}}&\makecell{47.47\\\small{\textcolor{green!50!black}{$\downarrow4.22$}}}&\makecell{49.82\\\small{\textcolor{red!75!black}{$\uparrow0.72$}}}&\makecell{49.12\\\small{\textcolor{green!50!black}{$\downarrow1.06$}}}&\makecell{51.95\\\small{\textcolor{green!50!black}{$\downarrow2.48$}}}&\makecell{21.88\\\small{\textcolor{green!50!black}{$\downarrow4.34$}}}&\makecell{25.38\\\small{\textcolor{green!50!black}{$\downarrow1.15$}}}&\makecell{21.24\\\small{\textcolor{green!50!black}{$\downarrow0.21$}}}&\makecell{21.33\\\small{\textcolor{green!50!black}{$\downarrow0.20$}}}&\makecell{23.40\\\small{\textcolor{green!50!black}{$\downarrow2.60$}}}&\makecell{26.60\\\small{\textcolor{green!50!black}{$\downarrow1.00$}}}&\makecell{50.00\\\small{\textcolor{green!50!black}{$\downarrow1.25$}}}\\
% \midrule
% Diff&\\
\bottomrule
\bottomrule
\end{tabular}
}
\setlength{\abovecaptionskip}{5pt}
\setlength{\belowcaptionskip}{-10pt}
\caption{Performance of combining weight sharing and quantization on Llama2-7B model.}
\label{tab:quant}
\end{table*}

\noindent
\textbf{Ablation on Head-wise Matching Functions.} 
For weight sharing, the choice of shared heads is critical. 
In Figure~\ref{fig:sharing_strategy}, we plot the performance curve on PIQA~\citep{bisk2019piqa} and OpenBookQA using different head-wise match functions for Baichuan2-7B model. 
And the corresponding detailed results are presented in Appendix~\ref{app:ablation}. 
Notably, using the cosine similarity between the concatenation matrix of $W^q$ and $W^k$ attains the most favorable outcomes. 
This may be because it guarantees the maximum similarities between attention maps from the model before and after weight sharing.
Also, this choice is much more stable and robust in some tasks like reading comprehension(e.g., OpenBookQA). 

\noindent
\textbf{Robustness on the Model Size.} 
In previous experiments, we adopt our approach in LLMs. 
Since small-scale models are not highly over-parameterized as large-scale  models~\citep{gao2023small}, we further verify the effectiveness of our method on smaller models like BERT-base, GPT2-small. 
For analysis, we set the sharing ratio from 0\% to 50\% with a step of 10\% for the fine-tuned GPT-small model on WMT-14 En-Fr dataset. 
As shown in Table~\ref{tab:gpt2-small}, at a 50\% sharing ratio, the GPT-small can still yield a BLEU score of 39.44 without any post-training. 
Such kind of variance in performance is acceptable that to some degree proves our method is also suitable for small-scale models. 

\begin{table}[htbp]
    \centering
    \setlength\tabcolsep{3pt} 
\large
    \resizebox{\linewidth}{!}{\begin{tabular}{c|cccccc}
    \toprule
    \toprule
    \textbf{Sharing Ratio}& 0\% & 10\% & 20\% & 30\%& 40\% & 50\%\\%&\textbf{\makecell{GPT2-small\\(124M)}}
    \midrule
    \textbf{BLEU}&\textbf{43.62}&42.49&41.95 &41.34&39.96&39.44\\
    % \midrule
    \textbf{Meteor}&\textbf{42.33}&40.75 &40.18 &38.43&37.21&36.62\\
    \bottomrule
    \bottomrule
    \end{tabular}
    }
    \setlength{\abovecaptionskip}{5pt}
    \setlength{\belowcaptionskip}{0pt}
    \caption{Robustness on the model size via PostShare (performed on GPT2-small using WMT-14 En-Fr).}
    \label{tab:gpt2-small}
    \vspace{-1.0em}
\end{table}

\noindent
\textbf{Combine Weight Sharing with Quantization.}
%With the increasing popularity of post-quantization, techniques like GPTQ~\citep{frantar2022gptq} and LLM.int8()~\citep{dettmers2022llm} have emerged to reduce the resource consumption of LLMs. 
In terms of saving memory, post-quantization employs the strategy of reducing precision in the LLM parameters, while weight sharing aims to reduce the number of parameters. 
From these two different directions, we suppose integrating weight sharing and quantization may help towards even more memory reduction of LLMs. 
Hence, we choose GPTQ~\citep{frantar2022gptq} as a representative and %run several benchmarks to 
test the effectiveness of applying two techniques in tandem. 
Specifically, 
we quantize Llama2-7B model after 30\% DirectShare to 4 bit precision. 
As reported in Table~\ref{tab:quant}, they can be effectively combined with no more than 5 points performance drop.

\begin{figure}[t]
    \centering
    % \vspace{-3pt}
    \includegraphics[width=0.88\linewidth]{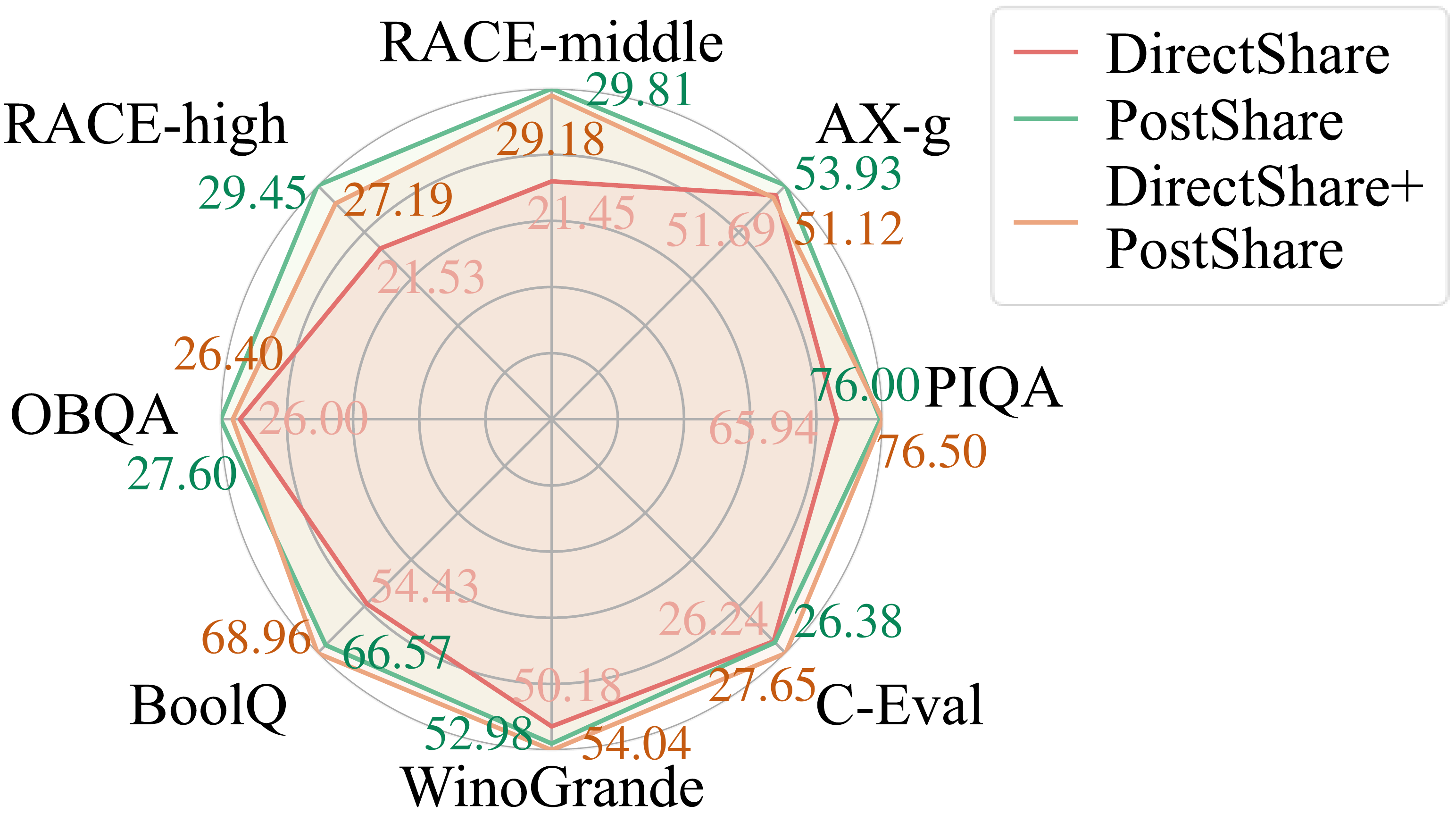}
    \setlength{\abovecaptionskip}{5pt}
    \setlength{\belowcaptionskip}{-10pt}
    \caption{Evaluation results when combining DirectShare with PostShare based on Llama2-7B model.}
    \label{fig:direct_post}
\end{figure}

\noindent
\textbf{Combine PostShare with DirectShare.} 
Another interesting research finding is the combination of our DirectShare and PostShare, where PostShare can play a role in fast performance recovery for DirectShare. 
Specifically, if we set the sharing ratio to 30\% and post-train only 0.5 epoch, the combination based on Llama2-7B performs on par with the PostShare, as Figure~\ref{fig:direct_post} shows. 
It can also be seen that DirectShare+PostShare outperforms in some specific datasets like BoolQ and WinoGrande, which we speculate is due to the mitigation of overfitting problem in PostShare to some extent. 
% We later confirm this hypothesis in Section~\ref{}. 

\noindent
\textbf{Visualization Study on the Shared Weights.} 
To provide a more detailed explanation of our rationale behind head-wise weight sharing, we conduct a visualization study on the ratios of weight sharing across the MHA layers in two models of different scales (see Appendix~\ref{app:visualization}). 
%The results , depicted in Figure~\ref{fig:match_distribution}, 
Results indicate the shareable weights distribution across attention heads is similar regardless of the sharing ratio.
We also observe a relative balanced sharing ratio across MHA layers than layer-wise weight sharing, which may seem counter-intuitive. 
However, we find such fine-grained operation on weights has already been used in model pruning~\cite{sun2023simple,ma2023llm}, constantly superior to layer-wise pruning. 

%% file: 6-Conclusion.tex
\section{Conclusion}
In this paper, we illustrate the feasibility of fine-grained weight sharing strategy applied in LLMs, namely, head-wise shareable attention. 
Consequently, we propose two methods for head-wise weight sharing called DirectShare and PostShare, which are complementary in terms of time and performance. 
Our DirectShare concentrates on a simple, no-training yet effective sharing strategy, performing competitively with one of the state-of-the-art model pruning methods. 
PostShare, on the other hand, shows an impressive performance on keeping LLM’s capabilities, needing to compromise on time efficiency. 
Last, we hope our work inspires researchers to explore better fine-grained weight sharing techniques for LLMs.

%% file: 7-Appendix.tex
\section{Detailed Explanation for DirectShare}\label{app:alg}
Algorithm~\ref{alg:sharing} summarizes the procedure of DirectShare:

\textbf{(1) Matching (line 1-13)}

We first initialize the buffer $\mathcal{D_\tau}$ \textbf{(line 1)} which is then used to store each candidate shareable attention head pairs. 
Next, the iterative process of matching begins: 

\textbf{Prepare Candidate Pairs (line 2-9)}:
Given the number of layers $\mathcal{L}$ and the number of heads per MHA block $\mathcal{H}$, we can construct candidate attention head pairs for sharing. 
For each pair, we then record the layer index and the head index accordingly, i.e., $index_i$ and $index_m$ respectively.

\textbf{Calculate Similarity (line 10-13)}:
We use 10 randomly selected samples from Wikipedia as the calibration samples for calculating the weight matrix similarity. 
We present the design of head-wise match function in Section~\ref{sec:strategy}. 
The final matching score $\mathcal{S}_{cos}$ is obtained by averaging across samples \textbf{(line 10)}. 
For every attention head, our approach selects the most matching counterpart according to $\mathcal{S}_{cos}$ \textbf{(line 11-12)} and stores them as pairs in the buffer $\mathcal{D_\tau}$ \textbf{(line 13)}.

\textbf{(2) Weight Sharing (line 14-16)}

We sort the candidate pairs in $\mathcal{D_\tau}$ in descending order of their matching scores $s_m$ \textbf{(line 14)}. 
Then, the top-N pairs $\mathcal{N}$ are selected according to the desired sharing ratio $\alpha$ \textbf{(line 15)}. 
Finally, for each selected head pair, we keep the weight values of the head in former layer unchanged and select to change the weight values of the head in latter layer \textbf{(line 16)}, resulting in the final model $\mathcal{M^*}$. 

\begin{algorithm2e}
% \SetAlgoLined % for end
\caption{\textbf{Weight\_Share} Function}
\SetInd{0.5em}{0.6em}
\KwIn{top-N matched head pairs $\mathcal{N}$, Original LLM $\mathcal{M}$}
\KwOut{The LLM $\mathcal{M^{*}}$ after weight sharing}
Initialize one dict $\mathcal{D}$\;
\For{pair in $\mathcal{N}$}{
    $index_i$, $index_j$ $\leftarrow pair$\;
    $(layer_i, i) \leftarrow index_i$\;
    $(layer_j, j) \leftarrow index_j$\;
    \eIf{$layer_i < layer_j$}{ 
        $\mathcal{D}[index_i]=index_j$\;
    }{
        $\mathcal{D}[index_j]=index_i$\;
    }
}
Sort $\mathcal{D}$ by the key\;
$\mathcal{M^{*}} \leftarrow \mathcal{M}$\;
\For{$head_i, head_j$ in $\mathcal{D}$}{
    $\mathcal{M^{*}} \leftarrow$ tie\_weight($\mathcal{M^{*}}$, $head_i$, $head_j$)\;
}
return $\mathcal{M}^*$\;
\end{algorithm2e}

\section{Relation between Matrix Weights and Attention Map Similarities}\label{app:similarity}
Our visual analysis (in Section\ref{sec:weight_similar}) can intuitively illustrate that the shared head pair with the highest similar weight matrices to a large degree exhibits highly similar attention maps across different datasets. 
Aside from this, we also provide a metric to measure the relation: 
\begin{equation}
    Degree = \frac{\#num~of~matched~heads}{\#num ~ of~ samples}
\end{equation}
Each time we use 100 randomly selected samples from each dataset to calculate the average cosine similarity value and repeat 5 times. 
According to the sharing ratio $\alpha$, we will respectively select the top-K shared head pairs based on the weight matrix similarity and attention map similarity (calculated cosine similarity as metric), denoted as Set $A$ and $B$.
The number of matched heads is calculated as the cardinality of the intersection of Set $A$ and Set $B$, denoted as $|A\cap B|$.
From Table~\ref{tab:similarity_measure}, it can be seen the matched degree is relatively high, so it is reasonable for us to do weight sharing directly across heads.

\begin{table}[htbp]
    \centering
    \resizebox{\linewidth}{!}{
    \begin{tabular}{c|c|c|c|c}
    \toprule
    \textbf{Dataset}	& WMT-14 & CQA & WSC & \textbf{Average}\\
    \midrule
    \textbf{Llama2-7B} &100\% &91.30\% &100\% &97.10\%\\
    \bottomrule
    \end{tabular}}
    \caption{The degree metric measured to quantify the relation between matrix weights and attention map similarities. CQA stands for CommonsenseQA benchmark. }
    \label{tab:similarity_measure}
    % \vspace{-10pt}
\end{table}

\section{Implementation Details}\label{app:experiments}
In this section, we will provide additional information about our experimental implementation. 

\subsection{Baselines}\label{app:baselines}
To our knowledge, there is no existing baseline for our methods, due to the absence of prior research on fine-grained weight sharing for LLMs. 
To provide a comprehensive demonstration of the effectiveness of our DirectShare, we can choose another important memory-efficient method of a different category for comparison. 
Here, we select two model pruning methods applied in LLMs: one classical model pruning method Magnitude Pruning and one state-of-the-art structured pruning method LLM-Pruner. 
We do not consider unstructured pruning methods in this paper since they can not achieve real memory reduction without specialized hardware or software. 
Appendix~\ref{app:more_comp} lists more comparison results with some baselines modified to support LLMs.

Based on the results presented in Table~\ref{tab:overall_results},~\ref{tab:b_reasoning},~\ref{tab:b_understanding},~\ref{tab:b_knowledge}, it is evident that our DirectShare performs on par with one of the prior best structured pruning methods regarding the overall performance, superior to the standard magnitude pruning. 
Consequently, we claim that designing such a fine-grained (i.e., head-wise) weight sharing strategy with a specific focus on LLMs is indeed simple but effective and this would be a good direction for future work. 

\subsection{Benchmarks}
\textbf{Logical and Common Sense Reasoning.} 
In the domain of reasoning, we consider two Chinese natural language inference benchmarks and three English benchmarks: CMNLI~\citep{xu2020clue}, OCNLI~\citep{hu2020ocnli}, 
%PIQA~\citep{bisk2019piqa}, 
along with AX-b, AX-g and RTE from SuperGLUE~\citep{wang2020superglue}. 

\noindent
\textbf{Natural Language Understanding (NLU).} 
In this field, we cover multiple tasks, including RACE~\citep{lai2017race} and OpenBookQA~\citep{OpenBookQA2018} for reading comprehension, CSL~\cite{li2022csl} for content summary and TNEWS~\citep{xu2020clue} for content analysis. 

\noindent
\textbf{Knowledge-related Tasks.} 
We perform evaluations regarding knowledge on various datasets: WinoGrande~\citep{levesque2012winograd} about language, BoolQ~\citep{clark2019boolq} testing knowledge question answering, C-Eval~\citep{huang2023ceval} and MMLU~\citep{hendryckstest2021} standing for two comprehensive examination benchmarks. 

\subsection{Post-training Details}\label{app:post_exp}
For carrying out the post-training process, we employ the code framework from LLaMA-Factory repository\footnote{https://github.com/hiyouga/LLaMA-Factory} with DeepSpeed ZeRO-1\footnote{Because of our designed special loss function in the post-training stage, only DeepSpeed ZeRO-1 can work.}. 
The Adam optimizer with a learning rate of 5e-5 is used in our experiment and the parameter values assigned during training are $\beta_1=0.9$ and $\beta_2=0.95$. 
For Llama 2-7B model, we set the batch size to 32. 
While for Llama 2-13B model, the batch size of training is only 8 subject to the limited computational resources. 
Besides, the maximum context size and $\gamma$ are set to 4096 and 0.5, respectively. 

\section{Experimental Results based on Baichuan 2 Models}\label{sec:baichuan2}
We re-implement Magnitude Pruning and LLM-Pruner with their public code to accommodate Baichuan2 models.
\begin{table}[!tbp]
\centering
\setlength\tabcolsep{3pt} 
\Large
\resizebox{\linewidth}{!}{
\begin{tabular}{@{}ccccm{1.4cm}m{1.4cm}m{1.4cm}@{}}
\toprule
\toprule
\textbf{Ratio} & \textbf{Method}
& \textbf{CMNLI} & \textbf{OCNLI} & \textbf{AX-b} & \textbf{AX-g} & \textbf{RTE} 
\\
\midrule
\textbf{0\%} & \textbf{Baichuan2-7B} &33.37&41.88&51.90&50.28&57.40\\
 \midrule
\multirow{3}{*}{\textbf{10\%}} &Magnitude&\underline{33.11}&33.12&\textbf{55.62}&\underline{50.84}&55.96\\
& LLM-Pruner&\textbf{37.31}&\underline{40.62}&49.18&50.00&\underline{60.65}\\
& DirectShare&33.00&\textbf{41.25}&\underline{49.55}&\textbf{51.12}&\underline{60.29}  \\
 \midrule
\multirow{3}{*}{\textbf{30\%}}
&Magnitude&\underline{32.97}&\underline{31.25}&\underline{48.28}&\textbf{51.97}&46.57\\
& LLM-Pruner&\textbf{34.20}&\textbf{34.38}&47.55&50.84&\textbf{51.26}\\
& DirectShare  & \underline{32.97} &30.63&\textbf{54.71}&\underline{51.69}&\underline{49.82}\\ 
\midrule
\midrule
\textbf{0\%} & \textbf{Baichuan2-13B} &33.21&40.62&59.69&50.59&44.77\\
 \midrule
\multirow{3}{*}{\textbf{10\%}}
&Magnitude&33.21&31.25& \underline{55.62}&48.60&46.93\\
& LLM-Pruner &\textbf{33.66}&\underline{36.88}&\textbf{58.51}&\underline{49.72}&\underline{47.65}\\
& DirectShare  &\underline{33.23}&\textbf{40.00}&53.71&\textbf{53.37}& \textbf{53.07}\\
 \midrule
\multirow{3}{*}{\textbf{30\%}}
&Magnitude&\textbf{33.21}&\underline{30.00}&50.91&48.03& 43.32\\
& LLM-Pruner &33.04&\textbf{36.88}&\textbf{55.71}&\textbf{50.28}&\underline{44.04} \\
& DirectShare &\underline{33.11}&\underline{30.00}&\underline{54.98}&\underline{50.00}&\textbf{45.13}\\ 
\bottomrule
\bottomrule
\end{tabular}
}
\vspace{-5pt}
\caption{Evaluation results on reasoning tasks when applying DirectShare to Baichuan2 models.}
\label{tab:b_reasoning}
\end{table}

\begin{table}[!tbp]
\centering
\setlength\tabcolsep{3pt} 
\Large
\resizebox{\linewidth}{!}{
\begin{tabular}{@{}ccccccc@{}}
\toprule
\toprule
\textbf{Ratio} & \textbf{Method}
& \textbf{\makecell{RACE- \\ middle}} & \textbf{\makecell{RACE- \\ high}} & 
\textbf{OBQA} %& \textbf{\makecell{OBQA- \\ fact}} 
& \textbf{CSL} & \textbf{TNEWS} 
\\
\midrule
\textbf{0\%} & \textbf{Baichuan2-7B} &51.04&52.63& 32.20&66.25&28.60 \\
 \midrule
\multirow{3}{*}{\textbf{10\%}}
&Magnitude&24.37&28.13&\underline{30.20}&57.50&\textbf{27.60}\\
& LLM-Pruner &\underline{25.42}&\underline{35.36}&\textbf{32.60}&\underline{61.25}&26.05 \\
& DirectShare &\textbf{50.49}&\textbf{48.46}&28.20&\textbf{63.75}&\underline{26.23}\\
 \midrule
\multirow{3}{*}{\textbf{30\%}}
&Magnitude&21.80&21.67&\textbf{27.60}&\textbf{57.50}&13.66\\
& LLM-Pruner&\underline{22.56}&\underline{22.67}&\underline{27.40}&\underline{53.12}&\textbf{21.31}\\
& DirectShare  &\textbf{25.14}&\textbf{23.44}&\textbf{27.60}&52.50&\underline{18.40}\\ 
\midrule
\midrule
\textbf{0\%} & \textbf{Baichuan2-13B} &68.94 &67.27 &42.20&63.12&28.96\\
 \midrule
\multirow{3}{*}{\textbf{10\%}}
&Magnitude&25.56&26.33&26.20&45.62&11.38\\
& LLM-Pruner & \underline{41.71}&\underline{46.80}&\textbf{32.40}&\underline{62.50}&\textbf{29.23}\\
& DirectShare  &\textbf{47.56} &\textbf{49.34}&\underline{31.20}&\textbf{64.38}&\underline{22.22}\\
 \midrule
\multirow{3}{*}{\textbf{30\%}}
&Magnitude&\textbf{24.58}&\textbf{24.58}&25.40&50.62&6.65\\
& LLM-Pruner &\underline{22.63}&21.81&\textbf{26.80}&\textbf{55.00}&\textbf{24.13}\\
& DirectShare &22.14&\underline{23.99}&\underline{26.60}&\underline{53.13}&\underline{17.58}\\ 
\bottomrule
\bottomrule
\end{tabular}
}
\vspace{-5pt}
\caption{NLU abilities of Baichuan2 models after DirectShare.}
\label{tab:b_understanding}
\vspace{-10pt}
\end{table}

\subsection{Logical and Common Sense Reasoning} \label{app:b_reasoning}
Table~\ref{tab:b_reasoning} presents a comparison on five datasets about reasoning abilities for three memory-efficient methods performed on the Baichuan2 models. 
Our results show that compared to NLU and knowledge-related abilities (listed in Table~\ref{tab:b_understanding},\ref{tab:b_knowledge}), DirectShare can indeed maintain its reasoning abilities to a large extent. 
Specifically, at 30\% ratio, DirectShare remains competitive with LLM-Pruner.

\subsection{Natural Language Understanding}\label{app:b_nlu}
Table~\ref{tab:b_understanding} presents the performance for each NLU task discussed in Section~\ref{sec:directshare} when applying DirectShare to Baichuan2 models. 
Consistent with the experiments on Llama2-7B and Llama2-13B models, similar performance drop exists. 
Thus, at the cost of post-training time, our PostShare can narrow the gap observed across the majority of datasets. 
With regard to individual datasets, it remains to be seen if the gap can be largely recovered given the best training epoch\footnote{We speculate that it may be attributed to overfitting issue. Furthermore, as the model size increases, it becomes increasingly difficult to determine the optimal training epoch for effectively mitigating overfitting.}.

\begin{table*}[!tbp]
\centering
\setlength\tabcolsep{3pt} 
\large
\resizebox{\textwidth}{!}{
\begin{tabular}{@{}cccccccccccc@{}}
\toprule
\toprule
\textbf{Ratio} & \textbf{Method}
& \textbf{WinoGrande}& \textbf{BoolQ} & \textbf{C-Eval} & \textbf{MMLU}&  \textbf{RACE-middle} & \textbf{RACE-high} & \textbf{OBQA}& \textbf{OBQA-fact}
\\
\midrule
\textbf{0\%} & \textbf{Llama 2-13B} &55.44&71.50&40.17&55.81&60.24&58.03& 42.40&60.00\\
\midrule
\multirow{2}{*}{\textbf{30\%}}& DirectShare &50.18&59.36&22.30&30.79&26.53&27.53&27.40&27.80\\
& PostShare$^*$ &53.68 \small{\textcolor{red}{$\uparrow3.50$}}&71.25 \small{\textcolor{red}{$\uparrow11.89$}}&25.80 \small{\textcolor{red}{$\uparrow3.50$}}&33.90 \small{\textcolor{red}{$\uparrow3.11$}}&32.03 \small{\textcolor{red}{$\uparrow3.30$}}&29.07 \small{\textcolor{red}{$\uparrow1.54$}}&33.60 \small{\textcolor{red}{$\uparrow6.20$}}&38.80 \small{\textcolor{red}{$\uparrow11.00$}}\\ 
\bottomrule
\bottomrule
\end{tabular}
}
\setlength{\abovecaptionskip}{5pt} 
\caption{Performance of PostShare based on the Llama2-13B backbone. * means choosing relatively good performance across different training steps.}
\label{tab:postshare_13b}
\vspace{-10pt}
\end{table*}

\begin{table}[!tbp]
\centering
\setlength\tabcolsep{3pt} 
\large
\resizebox{\linewidth}{!}{
\begin{tabular}{@{}cccccc@{}}
\toprule
\toprule
\textbf{Ratio} & \textbf{Method}
& \textbf{WinoGrande} & \textbf{BoolQ} & 
\textbf{C-Eval} & \textbf{MMLU} 
\\
\midrule
\textbf{0\%} & \textbf{Baichuan2-7B} &54.04 &63.30&56.19&54.65 \\
 \midrule
\multirow{3}{*}{\textbf{10\%}} 
&Magnitude&50.18&57.06&34.70&45.47\\
& LLM-Pruner &\underline{50.53}&\textbf{59.30}&\underline{48.14}&\textbf{51.78}  \\
& DirectShare  &\textbf{51.58}&\underline{58.01}&\textbf{50.41}&\underline{49.96}\\
 \midrule
\multirow{3}{*}{\textbf{30\%}}
&Magnitude&49.12&\textbf{55.41}&\textbf{23.91}&\underline{24.36} \\
& LLM-Pruner&\underline{51.23}&48.93&\underline{22.11}&\textbf{25.62} \\
& DirectShare  &\textbf{51.58}&\underline{51.53}&21.86&24.05\\ 
\midrule
\midrule
\textbf{0\%} & \textbf{Baichuan2-13B} &56.14&67.00&59.21&59.58\\
 \midrule
\multirow{3}{*}{\textbf{10\%}}
&Magnitude&50.53&40.55&25.22&25.55\\
& LLM-Pruner &\underline{51.23}&\textbf{65.87}&\underline{49.60}&\underline{51.49} \\
& DirectShare  & \textbf{53.33}&\underline{61.04}&\textbf{53.65}&\textbf{52.60}\\
 \midrule
\multirow{3}{*}{\textbf{30\%}}
&Magnitude&\underline{50.18}& \underline{50.09}&\textbf{25.35}&24.66\\
& LLM-Pruner&\textbf{50.53}&\textbf{59.42}&21.09&\textbf{24.95} \\
& DirectShare &48.77&40.83&\underline{23.25}&\underline{24.82}\\ 
\bottomrule
\bottomrule
\end{tabular}
}
\caption{Results on knowledge-related tasks of Baichuan2 models after DirectShare.}
\label{tab:b_knowledge}
\vspace{-10pt}
\end{table}

\subsection{Knowledge-related Tasks} \label{app:b_knowledge}
The results of Baichuan2 models on knowledge-related tasks are shown in Table~\ref{tab:b_knowledge}. 
Similar decline appears in Llama2-7B/13B models as well. 

\section{PostShare on Llama 2-13B Model}\label{app:postshare_13b}
In addition to Llama2-7B, we also experiment with Llama2-13B to evaluate PostShare (See Table~\ref{tab:postshare_13b}). 
Compared to Llama2-7B, the best training epoch on Llama2-13B is much smaller: approximately hundreds of training steps is enough, otherwise it may suffer from overfitting issue.
However, the overfitting problem seems to be obvious as model size increases, resulting in the challenge with regard to choosing the best training epoch. 

\begin{figure}[!tbp]
% \vspace{-1em}
    \centering
    \includegraphics[width=\linewidth]{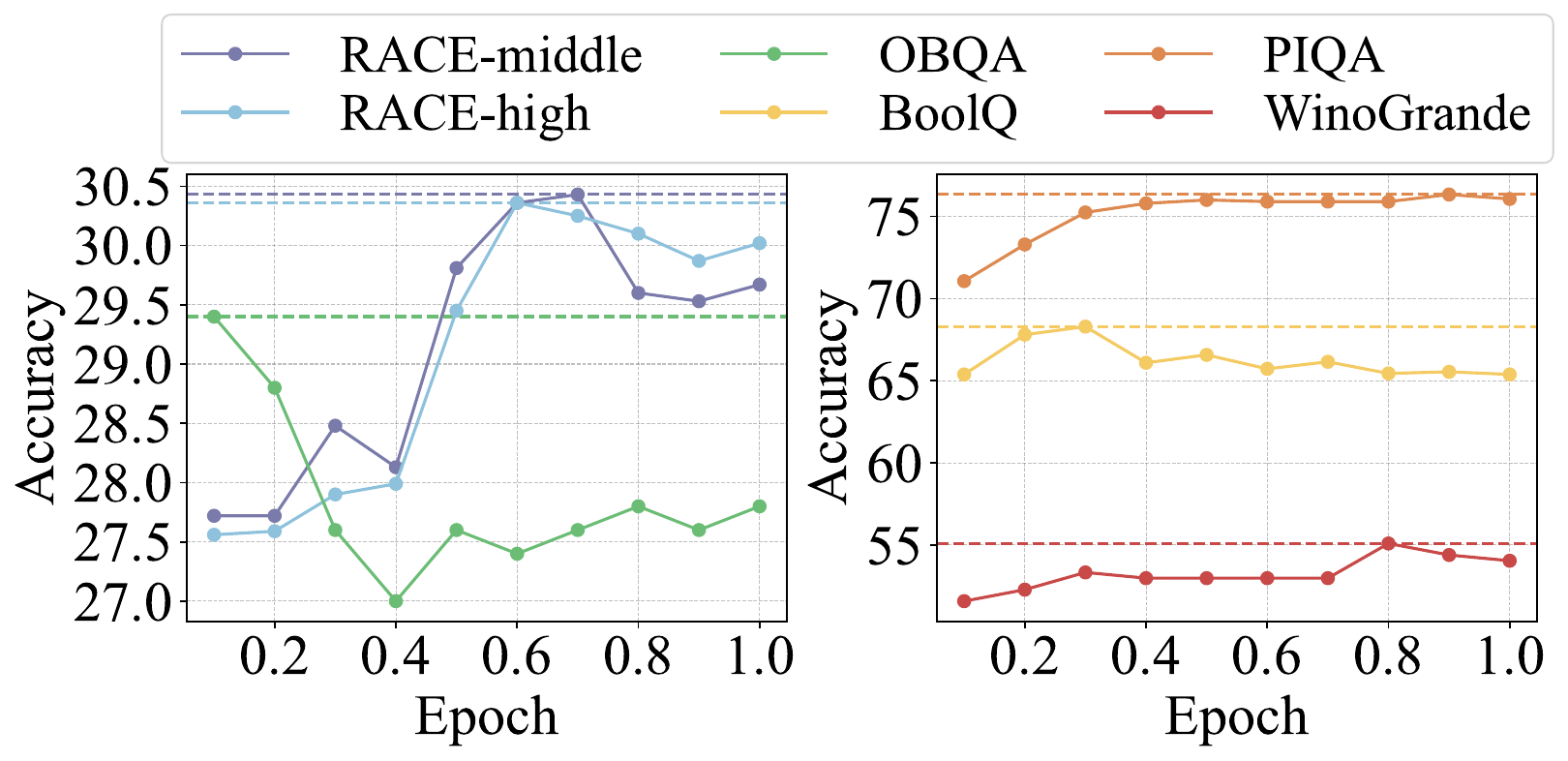}
    \setlength{\abovecaptionskip}{-10pt} 
    \setlength{\belowcaptionskip}{-5pt} 
    \caption{Accuracy across different training steps during PostShare.}
    \label{fig:overfitting}
\end{figure}

 \begin{table}[!tbp]
\centering
% \vspace{-1em}
\setlength\tabcolsep{3pt} 
\large
\resizebox{\linewidth}{!}{
\begin{tabular}{@{}ccccccc@{}}
\toprule
\toprule
\textbf{Epoch} &
   \textbf{\makecell{RACE-\\middle}} & \textbf{\makecell{RACE-\\high}} & \textbf{OBQA}%& \textbf{\makecell{OBQA-\\fact}}
   & \textbf{BoolQ} & \textbf{PIQA} & \textbf{\makecell{Wino-\\Grande}}
\\
\midrule
0.10&27.72&27.56&\textbf{29.40}&65.38&71.06&51.58\\
0.20&27.72&27.59&\underline{28.80}&\underline{67.80}&73.29&52.28\\ 
0.30&28.48&27.90&27.60&\textbf{68.29}&75.24&53.33\\ 
0.40&28.13&27.99&27.00&66.09&75.79&52.98\\ 
0.50&29.81&29.45&27.60&66.57&76.00&52.98\\ 
0.60&\underline{30.36}&\textbf{30.36}&27.40&65.72&75.90&52.98\\ 
0.70&\textbf{30.43}&\underline{30.25}&27.60&66.15&75.90&52.98\\ 
0.80&29.60&30.10&27.80&65.44&75.90&\textbf{55.09}\\ 
0.90&29.53&29.87&27.60&65.54&\textbf{76.33}&\underline{54.39}\\ 
1.00&29.67&30.02&27.80&65.38&\underline{76.06}&54.04\\ 
\bottomrule
\bottomrule
\end{tabular}
}
\vspace{-5pt}
\caption{Accuracy across different training steps during PostShare.}
\label{tab:overfitting}
\vspace{-15pt}
\end{table}

\section{More Analysis}
\subsection{Training Time Costs of PostShare}
For Llama2-7B model, our full-parameter post-training in English Wikipedia corpus uses around 11.27 hours for 0.5 epoch due to our limited computational resources. 
We list the post-training experimental settings in Appendix~\ref{app:post_exp}. 
In the same settings, the original full-parameter post-training requires 10.78 hours for 0.5 epoch, which indicates our PostShare does not increase much training costs (gap$\approx$30min).

\subsection{Overfitting Phenomenon in PostShare}
\label{app:overfitting}
% We present detailed statistical data in Table~\ref{tab:overfitting} showing the overfitting phenomenon in \textbf{PostShare}, as previously described in Section~\ref{sec:overfitting}. 
Figure~\ref{fig:overfitting} shows the performance curves on different kinds of datasets across various post-training steps. 
Remarkably, our PostShare requires no more than 1 epoch that can push the selected weights closer for sharing while keeping the performance. 
However, we observe the slight overfitting phenomenon in PostShare, i.e., the capabilities initially improve and then experience a slight decline. 
Besides, it is clear that the turning point about performance varies with datasets. 
Detailed statistical data are provided in Table~\ref{tab:overfitting}. 

\begin{table*}[htbp]
\centering
\setlength\tabcolsep{3pt} 
\large
\resizebox{\textwidth}{!}{
\begin{tabular}{@{}c|cc|cc|cc|cc|cc|cc|c|c@{}}
\toprule 	
\toprule 	
    \textbf{Sharing Ratio} & \multicolumn{2}{c}{\textbf{5\% }} & \multicolumn{2}{c}{\textbf{10\% }} & \multicolumn{2}{c}{\textbf{15\% }} & \multicolumn{2}{c}{\textbf{20\% }} & \multicolumn{2}{c}{\textbf{25\% }} & \multicolumn{2}{c}{\textbf{30\% }} & \multicolumn{1}{c}{\textbf{35\% }} & \multicolumn{1}{c}{\textbf{40\% }} \\
    \midrule
    \textbf{Dataset} & PIQA & OBQA&PIQA & OBQA&PIQA & OBQA&PIQA & OBQA&PIQA & OBQA&PIQA & OBQA&OBQA&OBQA\\
    \midrule
    $W^q$&\underline{74.92}&29.2&\underline{74.97}&27.5&73.29&\underline{27.8}&\underline{70.89}&27.7&64.64&\underline{27.5}&58.43&25.5&24.4&25.7\\
    $W^k$&\underline{74.92}&28.7&74.27&27.6&71.71&27.7&70.35&27.6&\underline{68.77}&\textbf{27.6}&\underline{64.36}&27.2&\underline{27.6}&\underline{26.9}\\
    $W^v$&\underline{74.92}&28.1&74.48&27.7&73.29&26.7&70.46&\textbf{28.5}&68.39&25.6&60.17&23.1&23.9&22.5\\
    $W^q,W^k,W^v$&71.71&27.6&63.55&27.8&54.03&26.8&50.16&24.5&51.41&25.5&51.09&25.5&\textbf{29.0}&25.3\\
    $W^q||W^k||W^v$&74.59&\textbf{34.7}&74.59&\textbf{30.0}&\underline{73.45}&\textbf{30.3}&70.73&\underline{28.2}&66.59&\textbf{27.6}&63.33&\underline{27.6}&27.1&25.0\\
    $W^q||W^k$(Ours)&\textbf{75.84}&\underline{33.9}&\textbf{75.30}&\underline{28.2}&\textbf{74.54}&27.5&\textbf{73.01}&27.3&\textbf{69.37}&\underline{27.5}&\textbf{65.56}&\textbf{28.0}&\underline{27.6}&\textbf{28.6}\\
\bottomrule
\bottomrule
\end{tabular}
}
\vspace{-5pt}
\caption{Results on PIQA and OBQA with different head-wise matching functions for Baichuan2-7B model.}
\label{tab:match_func}
\vspace{-5pt}
\end{table*}

\begin{table*}[!tbp]
\centering
\setlength\tabcolsep{3pt} 
\large
\resizebox{\textwidth}{!}{
\begin{tabular}{@{}c|cccccccccccccc|c@{}}
\toprule
\toprule
&CMNLI	&OCNLI	&AX-b	&AX-g	&RTE	&\makecell[c]{RACE-\\middle}	&\makecell[c]{RACE-\\high}	&OBQA	&CSL	&TNEWS	&\makecell[c]{Wino-\\Grande}	&BoolQ	&C-Eval	&MMLU	&Avg.\\
\midrule
Choosing A &33.33	&32.50&	57.07	&51.68	&49.10	&21.45	&21.53	&26.00	&51.25	&20.22	&50.18	&54.43	&26.24	&26.53	&\textbf{37.25}\\
Choosing B	&33.59	&31.25	&51.09	&53.09	&53.43	&25.00	&25.53	&22.20	&49.38	&12.20	&52.63	&57.22	&28.27	&25.63	&37.18\\
\bottomrule
\bottomrule
\end{tabular}}
\vspace{-5pt}
\caption{Performance comparison between choosing different shared weights for a pair of matched heads A, B. }
\label{tab:shared}
\end{table*}

\begin{figure*}[!tbp]
\centering
\vspace{-5pt}
\begin{subfigure}{\textwidth}
    \centering
    \includegraphics[width=\linewidth]{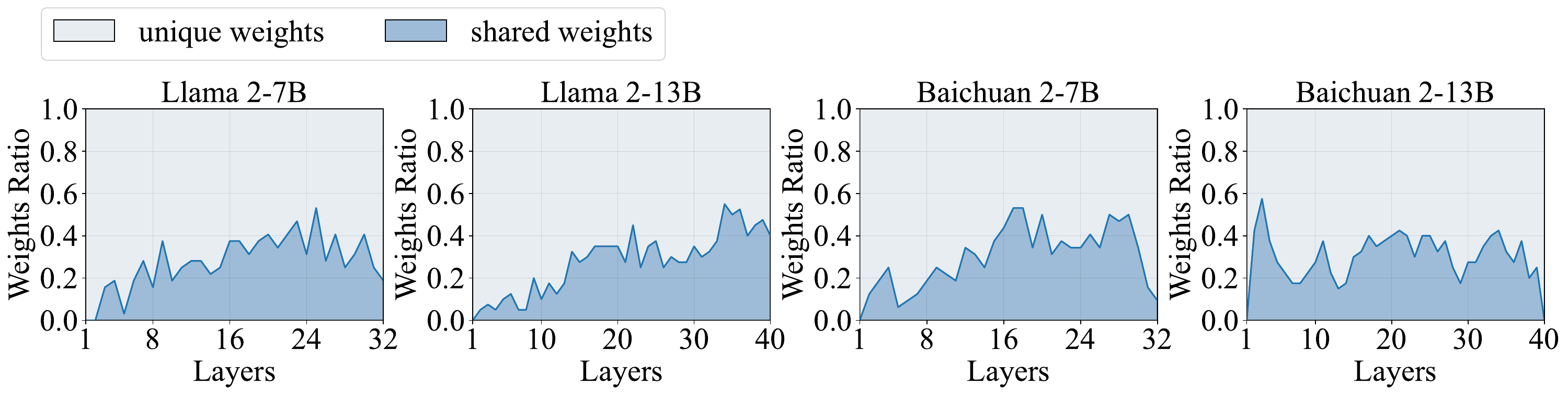}
    \setlength{\abovecaptionskip}{-13pt}
    \setlength{\belowcaptionskip}{1pt}
    \caption{Sharing Ratio=20\%}
\end{subfigure}
% \vspace{1.0em}
\begin{subfigure}{\textwidth}
    \centering
    \includegraphics[width=\linewidth]{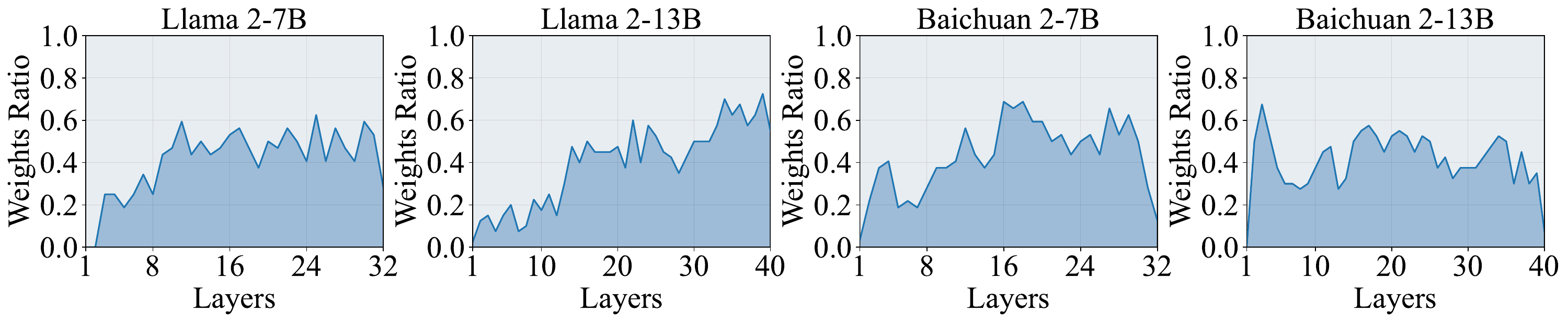}
    \setlength{\abovecaptionskip}{-13pt} 
    \caption{Sharing Ratio=30\%}
\end{subfigure}
\vspace{-20pt} 
    \caption{Ratios of weight sharing across the MHA Layers in Llama2-7B/13B and Baichuan2-7B/13B.}
    \label{fig:match_distribution}
\vspace{-10pt}
\end{figure*}

\vspace{-5pt}
\subsection{Impact of Different Head-wise Matching Functions}\label{app:ablation}
The selection of shared heads plays a crucial role in weight sharing. 
An ablation experiment for this is shown in Table~\ref{tab:match_func}. 

\vspace{-5pt}
\subsection{How to Choose the Shared Weights?}
We select the weight value from the head in the former layer as the shared weights. 
The reason behind is that we observe the weight values of the earlier layers are more sensitive in our experiments, which is also supported by previous studies~\citep{men2024shortgpt,chen2024compressing}. 
Thus, we tend to keep the weight values of the head in former layer unchanged.
We also additionally make a performance comparison between choosing different shared weights for a pair of matched heads A, B.  
Suppose head A is in the former layer and head B is in the latter layer. 
In Table~\ref{tab:shared}, it can be seen that the performance gap is small on the downstream tasks whether choosing A or B, but the overall performance is a little better when choosing the head weight in the former layers as the shared weights.

\subsection{Latency Evaluations}\label{app:latency}
Our paper aims to reduce the memory requirements for LLMs instead of accelerating. 
Since weight sharing does not change the total matrix multiplications, DirectShare can theoretically maintain the original model inference speed but our current implementation of head-wise weight sharing is hard to save memory and speed up at the same time, which may be a programming skill issue. 
Table~\ref{tab:latency} reports the inference speed tested under the test set of WikiText2. 
We will consider further optimization in future work to achieve a balance between memory and speedup.

\begin{table*}[!tbp]
\setlength\tabcolsep{3pt} 
\large
\resizebox{\textwidth}{!}{
\begin{tabular}{@{}cc|ccccccccccccccc@{}}
\toprule
\toprule
    Method&&CMNLI	&OCNLI	&AX-b	&AX-g	&RTE	&\makecell[c]{RACE-\\middle}	&\makecell[c]{RACE-\\high}	&OBQA	&CSL	&TNEWS	&\makecell[c]{Wino-\\Grande}	&BoolQ	&C-Eval	&MMLU	\\
    \midrule
    DirectShare	&Ratio=10\%	&33.00	&32.50	&54.17	&51.97	&50.90	&28.34	&28.96	&28.20	&54.37	&20.86	&52.63	&67.74	&28.75	&43.43\\
    \midrule
    GQA	&group size=2	&32.20	&27.50	&58.33	&50.00	&47.29	&25.21	&25.61	&22.20	&46.25	&6.92	&50.53	&37.83	&25.68	&25.80\\
    \midrule
    DirectShare	&Ratio=30\%	&33.33	&32.50	&57.07	&51.68	&49.10	&21.45	&21.53	&26.00	&51.25	&20.22	&50.18	&54.43	&26.24	&26.53\\
\bottomrule
\bottomrule
    \end{tabular}}
    \caption{Comparison results with GQA across various datasets.}
    \label{tab:cmp_gqa}
\end{table*}

\begin{table*}[!tbp]
\setlength\tabcolsep{3pt} 
\large
\resizebox{\textwidth}{!}{
\begin{tabular}{@{}cc|ccccccccccccccc@{}}
\toprule
\toprule
    &Ratio&CMNLI	&OCNLI	&AX-b	&AX-g	&RTE	&\makecell[c]{RACE-\\middle}	&\makecell[c]{RACE-\\high}	&OBQA	&CSL	&TNEWS	&\makecell[c]{Wino-\\Grande}	&BoolQ	&C-Eval	&MMLU	\\
    \midrule
Llama2-7B	&0\%	&32.98	&33.12	&53.53	&55.34	&49.82	&33.15	&35.51	&31.80	&55.62	&20.22	&54.04	&70.67	&32.20	&46.69\\
\midrule
DirectShare	&10\%	&33.00	&32.50	&54.17	&51.97	&50.90	&28.34	&28.96	&28.20	&54.37	&20.86	&52.63	&67.74	&28.75	&43.43\\
SEQ	&10\%	&31.55	&37.50	&49.82	&51.40	&52.35	&28.62	&28.39	&27.00	&57.50	&15.39	&51.23	&64.77	&30.40	&39.69\\
\midrule
DirectShare	&30\%	&33.33	&32.50	&57.07	&51.68	&49.10	&21.45	&21.53	&26.00	&51.25	&20.22	&50.18	&54.43	&26.24	&26.53\\
SEQ	&30\%	&31.72	&32.50	&47.55	&50.00	&46.93	&22.63	&22.30	&26.20	&50.62	&9.83	&49.47	&52.60	&22.05	&23.49\\
\bottomrule
\bottomrule
    \end{tabular}}
    \caption{Comparison results with SEQ across various datasets.}
    \label{tab:cmp_seq}
    \vspace{-10pt}
\end{table*}
\begin{table}[htbp]
\centering
\resizebox{0.9\linewidth}{!}{
\begin{tabular}{c|c|c}
    \toprule
    \toprule
    Model	&Sharing Ratio	&Speed\\
    \midrule
    Llama2-7B	&0\%	&41.70 tokens/s\\
    &30\% MHA	& 8.52 tokens/s\\
    &30\% MHA+FFN	& 8.34 tokens/s\\
    \midrule
    Llama2-13B	&	0\%	&24.20 tokens/s\\
    &30\% MHA & 5.63 tokens/s\\
    &30\% MHA+FFN & 5.67 tokens/s\\
    \bottomrule
    \bottomrule
    \end{tabular}}
    \caption{The inference speed of our method tested on a single A800-80GB GPU.}
    \label{tab:latency}
    \vspace{-10pt}
\end{table}

\subsection{Visualization Study on the Shared Weights}\label{app:visualization}
As depicted in Figure~\ref{fig:match_distribution}, the distribution of ratios of shared weights across attention heads is similar regardless of the sharing ratio.

\section{More Comparison Results}\label{app:more_comp}
\subsection{Compared with Grouped Query Attention}
Grouped Query Attention (GQA) has been widely adopted in LLMs to reduce the number of attention computations required, by grouping the query vectors into a smaller number of groups. 
Although GQA might be less flexible in choosing the reduction ratio of parameters, we also provide a performance comparison with our method. 
To equip the original Llama2-7B model with GQA, we perform the standard conversion proposed by ~\citet{ainslie2023gqa} and uses neighbor grouping with group size=2 to merge heads. 
The performance comparison is shown in Table~\ref{tab:cmp_gqa}. 
Furthermore, we compute the average scores per category (Reasoning, NLU, Knowledge) along with the average scores across all benchmarks (Avg.) to provide a more intuitive comparison in Table~\ref{tab:cmp_gqa_1}. 
Overall, our method demonstrates superior performance over GQA.

\begin{table}[htbp]
    \centering
    \setlength\tabcolsep{3pt} 
    \large
    \resizebox{\linewidth}{!}{
    \begin{tabular}{cc|ccc|c}
    \toprule
    \toprule
    &&Reasoning	&NLU	&Knowledge	&Avg.\\
    \midrule
    DirectShare	&Ratio=10\%	&44.51	&32.15	&48.14	&41.60\\
    \midrule
    GQA	&group size=2	&37.06	&25.24	&34.96	&32.42\\
    \midrule
    DirectShare	&Ratio=30\%	&44.74	&28.09	&39.35&	37.39\\
    \bottomrule
    \bottomrule
    \end{tabular}}
    \caption{Average comparison results with GQA.}
    \label{tab:cmp_gqa_1}
\end{table}

\subsection{Compared with Modified Layer-wise Weight Sharing}
We state the reason why choosing model pruning works as baselines in Section~\ref{app:baselines}. 
Prior layer-wise weight sharing techniques are proposed to apply to small-scale models like BERT. 
Considering one representative layer-wise weight sharing method~\cite{takase-kiyono-2021-rethinking}, we need to modify it to support LLMs, denoted as SEQ. 
For fair comparison, we perform layer-wise weight sharing in latter 60\% layers instead of all layers. 
The overall performance comparison based on Llama2-7B model is presented in Table~\ref{tab:cmp_seq}. 
Moreover, we calculate the overall average scores across all benchmarks (Avg.), the average scores for each category (Reasoning, NLU, Knowledge), and the average performance percentages relative to the original model across all benchmarks (Per.). 
The results in Table~\ref{tab:cmp_seq_1} indicate that our head-wise weight sharing method performs better than layer-wise weight sharing.

\begin{table}[htbp]
    \centering
    \setlength\tabcolsep{3pt} 
    \large
    \resizebox{\linewidth}{!}{
    \begin{tabular}{cc|cc|ccc}
    \toprule
    \toprule
    &Ratio	&Avg.	&Per.	&Reasoning	&NLU	&Knowledge	\\
    \midrule
Llama2-7B	&0\%	&43.71&	100\%	&44.96	&35.26	&50.90\\
\midrule
DirectShare	&10\%	&41.60	&95.17\%	&44.51	&32.15	&48.14\\
SEQ	&10\%	&40.74	&93.21\%	&44.52	&31.18	&46.53\\
\midrule
DirectShare	&30\%	&37.39	&85.54\%	&44.74	&28.09	&39.35\\
SEQ	&30\%	&34.97	&80.00\%	&41.68	&26.32	&36.90\\
    \bottomrule
    \bottomrule
    \end{tabular}}
    \caption{Average comparison results with SEQ.}
    \label{tab:cmp_seq_1}
\end{table}

\section{Generations from Models after Weight Sharing}
Table~\ref{tab:shared_llama2_7b}%,~\ref{tab:}
, \ref{tab:shared_baichuan2_7b} and \ref{tab:gpt2_small_translation} show more examples of the models after head-wise weight sharing. 
We present the generation results of both the shared model via DirectShare and PostShare. 

\begin{table}[htbp]
    \centering
    \setlength\tabcolsep{3pt} 
    \large
    \resizebox{\linewidth}{!}{
    \Large
    \begin{tabular}{p{3cm}|p{10cm}}
    \toprule
    \toprule
    \textbf{Model } &   \textbf{Generated Responses} \\
    \midrule
    \multirow{13}{3cm}{Llama2-7B \\w/ DirectShare} & \underline{The future is going to be one that presents many} \underline{challenges,} but also a great deal of opportunity. Everybody knows this, and it’s why we have been working so hard on the newest version of the future. The problem with the future is that it’s not always easy to see what you are doing wrong. It can be difficult to find \\
    \cmidrule{2-2}
    &\underline{Neural Sequential Model, especially transformers,} are a powerful tool for the analysis of time-series data. surely, they can be used to predict the future. Their use is not limited to the financial industry. In this article, we will discuss the benefits and limitations of the \\
    % \cmidrule{2-2}
    % &yyy\\
  \midrule
     \multirow{13}{3cm}{Llama2-7B \\w/ PostShare} & \underline{The future is going to be one that presents many} \underline{challenges,}  and the world will need a new generation of leaders. The United Nations has identified the following as key issues for the future: Climate change, Food security, Energy security, Water scarcity, Global health, Human rights, International law, Peacekeeping, Sustainable\\
     \cmidrule{2-2}
    &\underline{Neural Sequential Model, especially transformers,} is a type of artificial neural network that uses a sequence of vectors to represent the input and output. The model was introduced by Geoffrey Hinton in 1987. It has been used for speech recognition, handwriting recognition,\\
    % \cmidrule{2-2}
    % &yyy\\
  \bottomrule
  \bottomrule
    \end{tabular}
    }
    \setlength{\belowcaptionskip}{-5pt} 
    \caption{Responses from Llama2-7B (Sharing \\Ratio=30\%).}
    \label{tab:shared_llama2_7b}
\end{table}

\begin{table}[htbp]
\vspace{-35pt}
    \centering
    \setlength\tabcolsep{3pt} 
    \large
    \resizebox{\linewidth}{!}{
    \Large
    \begin{tabular}{p{3cm}|p{10cm}}
    \toprule
    \toprule
    \textbf{Model } &   \textbf{Generated Responses} \\
    \midrule
    \multirow{14}{3.0cm}{Baichuan2-7B w/ DirectShare (English)} & \underline{There are many reasons to support why persever-} \underline{ance is of great importance.} The first and foremost, it's the only way you can get better at what ever your doing! It also gives us an opportunity to learn from our mistakes or \\
    \cmidrule{2-2}
    &\underline{Her eyes shone with happiness.} “I’m so happy to see you,” She said, “and I am very proud of the work that we have done together in this project and it is a great honor for me as an artist because there are many people who has been working on these projects over the years but not all of them were able to do what was needed from us at any given time when things got out of hand or if they had lost their way” \\
    % \cmidrule{2-2}
    % &yyy\\
  \midrule
     \multirow{8}{3.0cm}{Baichuan2-7B w/ DirectShare (Chinese)} & \begin{CJK*}{UTF8}{gbsn}\underline{如何健康的生活} 1. 注意休息,保证睡眠。2.合理饮食:多吃水果、蔬菜等。3.适当运动(如散步) 4.保持乐观心态 5.多饮水 6.避免劳累7.不要熬夜 8.劳逸结合 9.注意安全 10.尊重生命\end{CJK*}\\
     \cmidrule{2-2}
    &\begin{CJK*}{UTF8}{gbsn}\underline{想要成功需要把握}什么?1. 首先,要有一个好的心态。2.其次,要有一定的经济基础3.接着就是自己的实力了4. 最后一个条件是:自己一定要努力!\end{CJK*}\\
    % \cmidrule{2-2}
    % &yyy\\
  \bottomrule
  \bottomrule
    \end{tabular}
    }
    \setlength{\belowcaptionskip}{-5pt} 
    \caption{Responses from Baichuan2-7B (Sharing \\Ratio=30\%).}
    \label{tab:shared_baichuan2_7b}
\end{table}

\begin{table}[htbp]
    \centering
    \vspace{20pt}
    \setlength\tabcolsep{3pt} 
    \LARGE
    \resizebox{\linewidth}{!}{
    \begin{tabular}{p{3cm}|p{7.9cm}|p{5.9cm}}
    \toprule
    \toprule
    \textbf{Model } &   \textbf{Generated Responses} &\textbf{Reference}\\
    \midrule
    \multirow{24}{3cm}{GPT2-small w/ PostShare} & \underline{In the letter, Mr Snowden said} \underline{he believed the support of the} \underline{international community could} \underline{persuade the US government to} \underline{abandon criminal charges again-} \underline{st him.<to-fr>} Dans la lettre, M. le député a dit que l'appui de la communauté internationale pourrait convaincre le gouvernement américain de abandonner la charge pénale contre lui. 
    & Dans sa lettre, M. Snowden a écrit qu'il pensait que le soutien de la communauté internationale pourrait persuader le gouvernement américain d'abandonner les charges pénales retenues contre lui. \\
    \cmidrule{2-3}
    &\underline{The letter extends an offer to co-} \underline{operate with German authorities} \underline{"when the difficulties of this hu-} \underline{manitarian situation have been} \underline{resolved."<to-fr>} La lettre élargit une invitation à coopérer avec les autorités allemands lorsque les difficultés de cette situation humanitaire sont réglées. & La lettre avance une offre de coopération avec les autorités allemandes « lorsque les difficultés de cette situation humanitaire auront été résolues ». \\
    \cmidrule{2-3}
    & \underline{The first test plane was unveiled} \underline{in March and took flight for the} \underline{first time in September after mo-} \underline{nths of delays.<to-fr>} Le premier étudiant a été démontré en mars et a fait l'avion pour la première fois après des mois de retard. & Le premier avion d'essai a été dévoilé en mars et s'est envolé pour la première fois en septembre après des mois de retard.\\
  \bottomrule
  \bottomrule
    \end{tabular}
    }
    \caption{Responses from GPT2-small (Sharing \\Ratio=30\%).}
    \label{tab:gpt2_small_translation}
\end{table}

%% file: main.bbl
\begin{thebibliography}{62}
\expandafter\ifx\csname natexlab\endcsname\relax\def\natexlab#1{#1}\fi

\bibitem[{Ainslie et~al.(2023)Ainslie, Lee-Thorp, de~Jong, Zemlyanskiy, Lebron, and Sanghai}]{ainslie2023gqa}
Joshua Ainslie, James Lee-Thorp, Michiel de~Jong, Yury Zemlyanskiy, Federico Lebron, and Sumit Sanghai. 2023.
\newblock Gqa: Training generalized multi-query transformer models from multi-head checkpoints.
\newblock In \emph{Proceedings of the 2023 Conference on Empirical Methods in Natural Language Processing}.

\bibitem[{Bai et~al.(2021)Bai, Zhang, Hou, Shang, Jin, Jiang, Liu, Lyu, and King}]{bai2020binarybert}
Haoli Bai, Wei Zhang, Lu~Hou, Lifeng Shang, Jin Jin, Xin Jiang, Qun Liu, Michael Lyu, and Irwin King. 2021.
\newblock {B}inary{BERT}: Pushing the limit of {BERT} quantization.
\newblock In \emph{Proceedings of the 59th Annual Meeting of the Association for Computational Linguistics and the 11th International Joint Conference on Natural Language Processing}, pages 4334--4348.

\bibitem[{Baichuan(2023)}]{baichuan2023baichuan2}
Baichuan. 2023.
\newblock Baichuan 2: Open large-scale language models.
\newblock \emph{ArXiv preprint}, abs/2309.10305.

\bibitem[{Bhojanapalli et~al.(2021)Bhojanapalli, Chakrabarti, Veit, Lukasik, Jain, Liu, Chang, and Kumar}]{bhojanapalli2021leveraging}
Srinadh Bhojanapalli, Ayan Chakrabarti, Andreas Veit, Michal Lukasik, Himanshu Jain, Frederick Liu, Yin-Wen Chang, and Sanjiv Kumar. 2021.
\newblock Leveraging redundancy in attention with reuse transformers.
\newblock \emph{ArXiv preprint}, abs/2110.06821.

\bibitem[{Bisk et~al.(2020)Bisk, Zellers, Gao, Choi et~al.}]{bisk2019piqa}
Yonatan Bisk, Rowan Zellers, Jianfeng Gao, Yejin Choi, et~al. 2020.
\newblock Piqa: Reasoning about physical commonsense in natural language.
\newblock In \emph{Proceedings of the AAAI conference on artificial intelligence}, volume~34, pages 7432--7439.

\bibitem[{Brown et~al.(2020)Brown, Mann, Ryder, Subbiah, Kaplan, Dhariwal, Neelakantan, Shyam, Sastry, Askell et~al.}]{brown2020language}
Tom Brown, Benjamin Mann, Nick Ryder, Melanie Subbiah, Jared~D Kaplan, Prafulla Dhariwal, Arvind Neelakantan, Pranav Shyam, Girish Sastry, Amanda Askell, et~al. 2020.
\newblock Language models are few-shot learners.
\newblock \emph{Advances in neural information processing systems}, 33:1877--1901.

\bibitem[{Bubeck et~al.(2023)Bubeck, Chandrasekaran, Eldan, Gehrke, Horvitz, Kamar, Lee, Lee, Li, Lundberg et~al.}]{bubeck2023sparks}
S{\'e}bastien Bubeck, Varun Chandrasekaran, Ronen Eldan, Johannes Gehrke, Eric Horvitz, Ece Kamar, Peter Lee, Yin~Tat Lee, Yuanzhi Li, Scott Lundberg, et~al. 2023.
\newblock Sparks of artificial general intelligence: Early experiments with gpt-4.
\newblock \emph{ArXiv preprint}, abs/2303.12712.

\bibitem[{Callison-Burch et~al.(2009)Callison-Burch, Koehn, Monz, and Schroeder}]{bojar-EtAl:2014:W14-33}
Chris Callison-Burch, Philipp Koehn, Christof Monz, and Josh Schroeder. 2009.
\newblock Findings of the 2009 {W}orkshop on {S}tatistical {M}achine {T}ranslation.
\newblock In \emph{Proceedings of the Fourth Workshop on Statistical Machine Translation}, pages 1--28.

\bibitem[{Chen et~al.(2024)Chen, Hu, and Zhang}]{chen2024compressing}
Xiaodong Chen, Yuxuan Hu, and Jing Zhang. 2024.
\newblock Compressing large language models by streamlining the unimportant layer.
\newblock \emph{ArXiv preprint}, abs/2403.19135.

\bibitem[{Chi et~al.(2021)Chi, Chung, Wu, Hsieh, Chen, Li, and Lee}]{chi2021audio}
Po-Han Chi, Pei-Hung Chung, Tsung-Han Wu, Chun-Cheng Hsieh, Yen-Hao Chen, Shang-Wen Li, and Hung-yi Lee. 2021.
\newblock Audio albert: A lite bert for self-supervised learning of audio representation.
\newblock In \emph{2021 IEEE Spoken Language Technology Workshop (SLT)}, pages 344--350. IEEE.

\bibitem[{Clark et~al.(2019)Clark, Lee, Chang, Kwiatkowski, Collins, and Toutanova}]{clark2019boolq}
Christopher Clark, Kenton Lee, Ming-Wei Chang, Tom Kwiatkowski, Michael Collins, and Kristina Toutanova. 2019.
\newblock Boolq: Exploring the surprising difficulty of natural yes/no questions.
\newblock \emph{arXiv preprint arXiv:1905.10044}.

\bibitem[{Contributors(2023)}]{2023opencompass}
OpenCompass Contributors. 2023.
\newblock Opencompass: A universal evaluation platform for foundation models.
\newblock \url{https://github.com/open-compass/opencompass}.

\bibitem[{Dabre and Fujita(2019)}]{dabre2019recurrent}
Raj Dabre and Atsushi Fujita. 2019.
\newblock Recurrent stacking of layers for compact neural machine translation models.
\newblock In \emph{Proceedings of the AAAI Conference on Artificial Intelligence}, volume~33, pages 6292--6299.

\bibitem[{Dehghani et~al.(2019)Dehghani, Gouws, Vinyals, Uszkoreit, and Kaiser}]{dehghani2018universal}
Mostafa Dehghani, Stephan Gouws, Oriol Vinyals, Jakob Uszkoreit, and Lukasz Kaiser. 2019.
\newblock Universal transformers.
\newblock In \emph{7th International Conference on Learning Representations}.

\bibitem[{Dettmers et~al.(2023)Dettmers, Svirschevski, Egiazarian, Kuznedelev, Frantar, Ashkboos, Borzunov, Hoefler, and Alistarh}]{dettmers2023spqr}
Tim Dettmers, Ruslan Svirschevski, Vage Egiazarian, Denis Kuznedelev, Elias Frantar, Saleh Ashkboos, Alexander Borzunov, Torsten Hoefler, and Dan Alistarh. 2023.
\newblock Spqr: A sparse-quantized representation for near-lossless llm weight compression.
\newblock \emph{ArXiv preprint}, abs/2306.03078.

\bibitem[{Devlin et~al.(2019)Devlin, Chang, Lee, and Toutanova}]{devlin2018bert}
Jacob Devlin, Ming-Wei Chang, Kenton Lee, and Kristina Toutanova. 2019.
\newblock {BERT}: Pre-training of deep bidirectional transformers for language understanding.
\newblock In \emph{Proceedings of the 2019 Conference of the North {A}merican Chapter of the Association for Computational Linguistics: Human Language Technologies}, pages 4171--4186.

\bibitem[{Foundation()}]{wikidump}
Wikimedia Foundation.
\newblock Wikimedia downloads.

\bibitem[{Frantar and Alistarh(2023)}]{frantar2023sparsegpt}
Elias Frantar and Dan Alistarh. 2023.
\newblock Sparsegpt: Massive language models can be accurately pruned in one-shot.
\newblock In \emph{International Conference on Machine Learning}, pages 10323--10337. PMLR.

\bibitem[{Frantar et~al.(2022)Frantar, Ashkboos, Hoefler, and Alistarh}]{frantar2022gptq}
Elias Frantar, Saleh Ashkboos, Torsten Hoefler, and Dan Alistarh. 2022.
\newblock Gptq: Accurate post-training quantization for generative pre-trained transformers.
\newblock \emph{ArXiv preprint}, abs/2210.17323.

\bibitem[{Gao et~al.(2023)Gao, Zhou, Liu, Zhao, and Wen}]{gao2023small}
Ze-Feng Gao, Kun Zhou, Peiyu Liu, Wayne~Xin Zhao, and Ji-Rong Wen. 2023.
\newblock Small pre-trained language models can be fine-tuned as large models via over-parameterization.
\newblock In \emph{Proceedings of the 61st Annual Meeting of the Association for Computational Linguistics}, pages 3819--3834.

\bibitem[{Hendrycks et~al.(2021)Hendrycks, Burns, Basart, Zou, Mazeika, Song, and Steinhardt}]{hendryckstest2021}
Dan Hendrycks, Collin Burns, Steven Basart, Andy Zou, Mantas Mazeika, Dawn Song, and Jacob Steinhardt. 2021.
\newblock Measuring massive multitask language understanding.
\newblock In \emph{9th International Conference on Learning Representations}.

\bibitem[{Hu et~al.(2022)Hu, Shen, Wallis, Allen{-}Zhu, Li, Wang, Wang, and Chen}]{hu2021lora}
Edward~J. Hu, Yelong Shen, Phillip Wallis, Zeyuan Allen{-}Zhu, Yuanzhi Li, Shean Wang, Lu~Wang, and Weizhu Chen. 2022.
\newblock Lora: Low-rank adaptation of large language models.
\newblock In \emph{The Tenth International Conference on Learning Representations}.

\bibitem[{Hu et~al.(2020)Hu, Richardson, Xu, Li, K{\"u}bler, and Moss}]{hu2020ocnli}
Hai Hu, Kyle Richardson, Liang Xu, Lu~Li, Sandra K{\"u}bler, and Lawrence Moss. 2020.
\newblock {OCNLI}: {O}riginal {C}hinese {N}atural {L}anguage {I}nference.
\newblock In \emph{Findings of the Association for Computational Linguistics: EMNLP 2020}.

\bibitem[{Huang et~al.(2023)Huang, Bai, Zhu, Zhang, Zhang, Su, Liu, Lv, Zhang, Lei, Fu, Sun, and He}]{huang2023ceval}
Yuzhen Huang, Yuzhuo Bai, Zhihao Zhu, Junlei Zhang, Jinghan Zhang, Tangjun Su, Junteng Liu, Chuancheng Lv, Yikai Zhang, Jiayi Lei, Yao Fu, Maosong Sun, and Junxian He. 2023.
\newblock C-eval: A multi-level multi-discipline chinese evaluation suite for foundation models.
\newblock \emph{ArXiv preprint}, abs/2305.08322.

\bibitem[{Kaplan et~al.(2020)Kaplan, McCandlish, Henighan, Brown, Chess, Child, Gray, Radford, Wu, and Amodei}]{kaplan2020scaling}
Jared Kaplan, Sam McCandlish, Tom Henighan, Tom~B Brown, Benjamin Chess, Rewon Child, Scott Gray, Alec Radford, Jeffrey Wu, and Dario Amodei. 2020.
\newblock Scaling laws for neural language models.
\newblock \emph{ArXiv preprint}, abs/2001.08361.

\bibitem[{Kitaev et~al.(2020)Kitaev, Kaiser, and Levskaya}]{kitaev2020reformer}
Nikita Kitaev, Lukasz Kaiser, and Anselm Levskaya. 2020.
\newblock Reformer: The efficient transformer.
\newblock In \emph{8th International Conference on Learning Representations}.

\bibitem[{Lai et~al.(2017)Lai, Xie, Liu, Yang, and Hovy}]{lai2017race}
Guokun Lai, Qizhe Xie, Hanxiao Liu, Yiming Yang, and Eduard Hovy. 2017.
\newblock {RACE}: Large-scale {R}e{A}ding comprehension dataset from examinations.
\newblock In \emph{Proceedings of the 2017 Conference on Empirical Methods in Natural Language Processing}.

\bibitem[{Lan et~al.(2020)Lan, Chen, Goodman, Gimpel, Sharma, and Soricut}]{lan2019albert}
Zhenzhong Lan, Mingda Chen, Sebastian Goodman, Kevin Gimpel, Piyush Sharma, and Radu Soricut. 2020.
\newblock {ALBERT:} {A} lite {BERT} for self-supervised learning of language representations.
\newblock In \emph{8th International Conference on Learning Representations}.

\bibitem[{Levesque et~al.(2012)Levesque, Davis, and Morgenstern}]{levesque2012winograd}
Hector Levesque, Ernest Davis, and Leora Morgenstern. 2012.
\newblock The winograd schema challenge.
\newblock In \emph{Thirteenth International Conference on the Principles of Knowledge Representation and Reasoning}. Citeseer.

\bibitem[{Li et~al.(2022)Li, Zhang, Zhao, Shen, Liu, Mao, and Zhang}]{li2022csl}
Yudong Li, Yuqing Zhang, Zhe Zhao, Linlin Shen, Weijie Liu, Weiquan Mao, and Hui Zhang. 2022.
\newblock {CSL}: A large-scale {C}hinese scientific literature dataset.
\newblock In \emph{Proceedings of the 29th International Conference on Computational Linguistics}, pages 3917--3923.

\bibitem[{Li et~al.(2020)Li, Wallace, Shen, Lin, Keutzer, Klein, and Gonzalez}]{li2020train}
Zhuohan Li, Eric Wallace, Sheng Shen, Kevin Lin, Kurt Keutzer, Dan Klein, and Joey Gonzalez. 2020.
\newblock Train big, then compress: Rethinking model size for efficient training and inference of transformers.
\newblock In \emph{International Conference on machine learning}, pages 5958--5968. PMLR.

\bibitem[{Liu et~al.(2023)Liu, Gao, Chen, Zhao, and Wen}]{liu2023enhancing}
Peiyu Liu, Ze-Feng Gao, Yushuo Chen, Wayne~Xin Zhao, and Ji-Rong Wen. 2023.
\newblock Enhancing scalability of pre-trained language models via efficient parameter sharing.
\newblock In \emph{Findings of the Association for Computational Linguistics: EMNLP 2023}.

\bibitem[{Liu et~al.(2019)Liu, Ott, Goyal, Du, Joshi, Chen, Levy, Lewis, Zettlemoyer, and Stoyanov}]{liu2019roberta}
Yinhan Liu, Myle Ott, Naman Goyal, Jingfei Du, Mandar Joshi, Danqi Chen, Omer Levy, Mike Lewis, Luke Zettlemoyer, and Veselin Stoyanov. 2019.
\newblock Roberta: A robustly optimized bert pretraining approach.
\newblock \emph{ArXiv preprint}, abs/1907.11692.

\bibitem[{Lv et~al.(2023)Lv, Zhang, Li, Gan, and Sun}]{lv2023lightformer}
Xiuqing Lv, Peng Zhang, Sunzhu Li, Guobing Gan, and Yueheng Sun. 2023.
\newblock Lightformer: Light-weight transformer using svd-based weight transfer and parameter sharing.
\newblock In \emph{Findings of the Association for Computational Linguistics: ACL 2023}.

\bibitem[{Ma et~al.(2023)Ma, Fang, and Wang}]{ma2023llm}
Xinyin Ma, Gongfan Fang, and Xinchao Wang. 2023.
\newblock Llm-pruner: On the structural pruning of large language models.
\newblock \emph{ArXiv preprint}, abs/2305.11627.

\bibitem[{Men et~al.(2024)Men, Xu, Zhang, Wang, Lin, Lu, Han, and Chen}]{men2024shortgpt}
Xin Men, Mingyu Xu, Qingyu Zhang, Bingning Wang, Hongyu Lin, Yaojie Lu, Xianpei Han, and Weipeng Chen. 2024.
\newblock Shortgpt: Layers in large language models are more redundant than you expect.
\newblock \emph{ArXiv preprint}, abs/2403.03853.

\bibitem[{Mihaylov et~al.(2018)Mihaylov, Clark, Khot, and Sabharwal}]{OpenBookQA2018}
Todor Mihaylov, Peter Clark, Tushar Khot, and Ashish Sabharwal. 2018.
\newblock Can a suit of armor conduct electricity? a new dataset for open book question answering.
\newblock In \emph{Proceedings of the 2018 Conference on Empirical Methods in Natural Language Processing}, pages 2381--2391.

\bibitem[{Reid et~al.(2021)Reid, Marrese-Taylor, and Matsuo}]{reid2021subformer}
Machel Reid, Edison Marrese-Taylor, and Yutaka Matsuo. 2021.
\newblock Subformer: Exploring weight sharing for parameter efficiency in generative transformers.
\newblock In \emph{Findings of the Association for Computational Linguistics: EMNLP 2021}.

\bibitem[{Shim et~al.(2023)Shim, Choi, and Sung}]{shim2023exploring}
Kyuhong Shim, Jungwook Choi, and Wonyong Sung. 2023.
\newblock Exploring attention map reuse for efficient transformer neural networks.
\newblock \emph{ArXiv preprint}, abs/2301.12444.

\bibitem[{Sun et~al.(2023)Sun, Liu, Bair, and Kolter}]{sun2023simple}
Mingjie Sun, Zhuang Liu, Anna Bair, and J~Zico Kolter. 2023.
\newblock A simple and effective pruning approach for large language models.
\newblock \emph{ArXiv preprint}, abs/2306.11695.

\bibitem[{Takase and Kiyono(2021{\natexlab{a}})}]{takase2021lessons}
Sho Takase and Shun Kiyono. 2021{\natexlab{a}}.
\newblock Lessons on parameter sharing across layers in transformers.
\newblock \emph{ArXiv preprint}, abs/2104.06022.

\bibitem[{Takase and Kiyono(2021{\natexlab{b}})}]{takase-kiyono-2021-rethinking}
Sho Takase and Shun Kiyono. 2021{\natexlab{b}}.
\newblock \href {https://doi.org/10.18653/v1/2021.naacl-main.460} {Rethinking perturbations in encoder-decoders for fast training}.
\newblock In \emph{Proceedings of the 2021 Conference of the North American Chapter of the Association for Computational Linguistics: Human Language Technologies}, pages 5767--5780, Online. Association for Computational Linguistics.

\bibitem[{Talmor et~al.(2019)Talmor, Herzig, Lourie, and Berant}]{talmor-etal-2019-commonsenseqa}
Alon Talmor, Jonathan Herzig, Nicholas Lourie, and Jonathan Berant. 2019.
\newblock \href {https://doi.org/10.18653/v1/N19-1421} {{C}ommonsense{QA}: A question answering challenge targeting commonsense knowledge}.
\newblock In \emph{Proceedings of the 2019 Conference of the North {A}merican Chapter of the Association for Computational Linguistics: Human Language Technologies, Volume 1 (Long and Short Papers)}, pages 4149--4158, Minneapolis, Minnesota. Association for Computational Linguistics.

\bibitem[{Tan et~al.(2023)Tan, Tam, Wang, Gong, Yang, Tang, He, Liu, Wang, Zhao et~al.}]{tan2023gkd}
Shicheng Tan, Weng~Lam Tam, Yuanchun Wang, Wenwen Gong, Yang Yang, Hongyin Tang, Keqing He, Jiahao Liu, Jingang Wang, Shu Zhao, et~al. 2023.
\newblock Gkd: A general knowledge distillation framework for large-scale pre-trained language model.
\newblock \emph{ArXiv preprint}, abs/2306.06629.

\bibitem[{Tao et~al.(2023)Tao, Hou, Bai, Wei, Jiang, Liu, Luo, and Wong}]{tao2023structured}
Chaofan Tao, Lu~Hou, Haoli Bai, Jiansheng Wei, Xin Jiang, Qun Liu, Ping Luo, and Ngai Wong. 2023.
\newblock Structured pruning for efficient generative pre-trained language models.
\newblock In \emph{Findings of the Association for Computational Linguistics: ACL 2023}.

\bibitem[{Tao et~al.(2022)Tao, Hou, Zhang, Shang, Jiang, Liu, Luo, and Wong}]{tao2022compression}
Chaofan Tao, Lu~Hou, Wei Zhang, Lifeng Shang, Xin Jiang, Qun Liu, Ping Luo, and Ngai Wong. 2022.
\newblock Compression of generative pre-trained language models via quantization.
\newblock In \emph{Proceedings of the 60th Annual Meeting of the Association for Computational Linguistics}, pages 4821--4836.

\bibitem[{Touvron et~al.(2023)Touvron, Martin, Stone, Albert, Almahairi, Babaei, Bashlykov, Batra, Bhargava, Bhosale et~al.}]{touvron2023llama}
Hugo Touvron, Louis Martin, Kevin Stone, Peter Albert, Amjad Almahairi, Yasmine Babaei, Nikolay Bashlykov, Soumya Batra, Prajjwal Bhargava, Shruti Bhosale, et~al. 2023.
\newblock Llama 2: Open foundation and fine-tuned chat models.
\newblock \emph{ArXiv preprint}, abs/2307.09288.

\bibitem[{Vig(2019)}]{vig2019multiscale}
Jesse Vig. 2019.
\newblock A multiscale visualization of attention in the transformer model.
\newblock In \emph{Proceedings of the 57th Annual Meeting of the Association for Computational Linguistics: System Demonstrations}, pages 37--42.

\bibitem[{Wang et~al.(2019)Wang, Pruksachatkun, Nangia, Singh, Michael, Hill, Levy, and Bowman}]{wang2020superglue}
Alex Wang, Yada Pruksachatkun, Nikita Nangia, Amanpreet Singh, Julian Michael, Felix Hill, Omer Levy, and Samuel Bowman. 2019.
\newblock Superglue: A stickier benchmark for general-purpose language understanding systems.
\newblock \emph{Advances in neural information processing systems}.

\bibitem[{Wei et~al.(2022)Wei, Tay, Bommasani, Raffel, Zoph, Borgeaud, Yogatama, Bosma, Zhou, Metzler et~al.}]{wei2022emergent}
Jason Wei, Yi~Tay, Rishi Bommasani, Colin Raffel, Barret Zoph, Sebastian Borgeaud, Dani Yogatama, Maarten Bosma, Denny Zhou, Donald Metzler, et~al. 2022.
\newblock Emergent abilities of large language models.
\newblock \emph{ArXiv preprint}, abs/2206.07682.

\bibitem[{Wu et~al.(2023)Wu, Chen, Quan, Wang, and Wang}]{wu2023ad}
Siyue Wu, Hongzhan Chen, Xiaojun Quan, Qifan Wang, and Rui Wang. 2023.
\newblock Ad-kd: Attribution-driven knowledge distillation for language model compression.
\newblock \emph{ArXiv preprint}, abs/2305.10010.

\bibitem[{Xia et~al.(2019)Xia, He, Tan, Tian, He, and Qin}]{xia2019tied}
Yingce Xia, Tianyu He, Xu~Tan, Fei Tian, Di~He, and Tao Qin. 2019.
\newblock Tied transformers: Neural machine translation with shared encoder and decoder.
\newblock In \emph{Proceedings of the AAAI conference on artificial intelligence}, volume~33, pages 5466--5473.

\bibitem[{Xiao et~al.(2019)Xiao, Li, Zhu, Yu, and Liu}]{xiao2019sharing}
Tong Xiao, Yinqiao Li, Jingbo Zhu, Zhengtao Yu, and Tongran Liu. 2019.
\newblock Sharing attention weights for fast transformer.
\newblock In \emph{Proceedings of the Twenty-Eighth International Joint Conference on Artificial Intelligence, {IJCAI} 2019}, pages 5292--5298.

\bibitem[{Xu and McAuley(2023)}]{xu2023compressionsurvey}
Canwen Xu and Julian McAuley. 2023.
\newblock A survey on model compression and acceleration for pretrained language models.
\newblock In \emph{Proceedings of the AAAI Conference on Artificial Intelligence}, pages 10566--10575.

\bibitem[{Xu et~al.(2020)Xu, Hu, Zhang, Li, Cao, Li, Xu, Sun, Yu, Yu, Tian, Dong, Liu, Shi, Cui, Li, Zeng, Wang, Xie, Li, Patterson, Tian, Zhang, Zhou, Liu, Zhao, Zhao, Yue, Zhang, Yang, Richardson, and Lan}]{xu2020clue}
Liang Xu, Hai Hu, Xuanwei Zhang, Lu~Li, Chenjie Cao, Yudong Li, Yechen Xu, Kai Sun, Dian Yu, Cong Yu, Yin Tian, Qianqian Dong, Weitang Liu, Bo~Shi, Yiming Cui, Junyi Li, Jun Zeng, Rongzhao Wang, Weijian Xie, Yanting Li, Yina Patterson, Zuoyu Tian, Yiwen Zhang, He~Zhou, Shaoweihua Liu, Zhe Zhao, Qipeng Zhao, Cong Yue, Xinrui Zhang, Zhengliang Yang, Kyle Richardson, and Zhenzhong Lan. 2020.
\newblock {CLUE}: A {C}hinese language understanding evaluation benchmark.
\newblock In \emph{Proceedings of the 28th International Conference on Computational Linguistics}, pages 4762--4772.

\bibitem[{Yang et~al.(2024)Yang, Cao, and Zhao}]{yang2024laco}
Yifei Yang, Zouying Cao, and Hai Zhao. 2024.
\newblock Laco: Large language model pruning via layer collapse.
\newblock \emph{ArXiv preprint}, abs/2402.11187.

\bibitem[{Yang et~al.(2022)Yang, Cui, Yao, and Wang}]{yang2022gradient}
Ziqing Yang, Yiming Cui, Xin Yao, and Shijin Wang. 2022.
\newblock Gradient-based intra-attention pruning on pre-trained language models.
\newblock \emph{ArXiv preprint}, abs/2212.07634.

\bibitem[{Ying et~al.(2021)Ying, Ke, He, and Liu}]{ying2021lazyformer}
Chengxuan Ying, Guolin Ke, Di~He, and Tie-Yan Liu. 2021.
\newblock Lazyformer: Self attention with lazy update.
\newblock \emph{ArXiv preprint}, abs/2102.12702.

\bibitem[{Zhang et~al.(2022)Zhang, Peng, Wu, Liu, Xiao, Fu, and Yuan}]{zhang2022minivit}
Jinnian Zhang, Houwen Peng, Kan Wu, Mengchen Liu, Bin Xiao, Jianlong Fu, and Lu~Yuan. 2022.
\newblock Minivit: Compressing vision transformers with weight multiplexing.
\newblock In \emph{Proceedings of the IEEE/CVF Conference on Computer Vision and Pattern Recognition}, pages 12145--12154.

\bibitem[{Zhao et~al.(2023)Zhao, Zhou, Li, Tang, Wang, Hou, Min, Zhang, Zhang, Dong et~al.}]{zhao2023survey}
Wayne~Xin Zhao, Kun Zhou, Junyi Li, Tianyi Tang, Xiaolei Wang, Yupeng Hou, Yingqian Min, Beichen Zhang, Junjie Zhang, Zican Dong, et~al. 2023.
\newblock A survey of large language models.
\newblock \emph{ArXiv preprint}, abs/2303.18223.

\bibitem[{Zhou et~al.(2021)Zhou, Kang, Jin, Yang, Lian, Jiang, Hou, and Feng}]{zhou2021deepvit}
Daquan Zhou, Bingyi Kang, Xiaojie Jin, Linjie Yang, Xiaochen Lian, Zihang Jiang, Qibin Hou, and Jiashi Feng. 2021.
\newblock Deepvit: Towards deeper vision transformer.
\newblock \emph{ArXiv preprint}, abs/2103.11886.

\bibitem[{Zhu and Gupta(2017)}]{zhu2017prune}
Michael Zhu and Suyog Gupta. 2017.
\newblock To prune, or not to prune: exploring the efficacy of pruning for model compression.
\newblock \emph{ArXiv preprint}, abs/1710.01878.

\end{thebibliography}
